\PassOptionsToPackage{table}{xcolor}
\documentclass{article}
\usepackage{iclr2026_conference,times}
\iclrfinalcopy  

\usepackage{amsmath,amsfonts,bm}

\def\eqref#1{equation~\ref{#1}}

\def\1{\bm{1}}

\DeclareMathAlphabet{\mathsfit}{\encodingdefault}{\sfdefault}{m}{sl}
\SetMathAlphabet{\mathsfit}{bold}{\encodingdefault}{\sfdefault}{bx}{n}

\usepackage{graphicx}
\usepackage{booktabs}
\usepackage{amsfonts}
\usepackage{nicefrac}
\usepackage{microtype}
\usepackage{amsmath}
\usepackage{algorithm}
\usepackage{algpseudocode}
\usepackage{float}
\usepackage{caption}
\usepackage{subcaption}
\usepackage{adjustbox}
\usepackage{multirow}
\usepackage{amssymb}
\usepackage{bm}
\usepackage{mathabx}
\usepackage[T1]{fontenc}
\usepackage{diagbox}
\usepackage[hidelinks]{hyperref}
\usepackage{url}
\definecolor{mylightgray}{gray}{0.9}
\title{MindPilot: Closed-loop Visual Stimulation Optimization for Brain Modulation with EEG-guided Diffusion}

\author{
\hspace{-0.5em}
\textbf{Dongyang Li}$^{1}$\thanks{Equal Contribution.}  \hspace{1.5em}
\textbf{Kunpeng Xie}$^{1}$\footnotemark[1]  \hspace{1.5em}
\textbf{Mingyang Wu}$^{1}$  \hspace{1.5em}
\textbf{Yiwei Kong}$^{1,2}$  \hspace{1.5em}
\textbf{Jiahua Tang}$^{1,3}$ 
\vspace{-2mm} \AND \vspace{2mm} 
\hspace{7.5em} 
\textbf{Haoyang Qin}$^{1}$ \hspace{1.5em}
\textbf{Chen Wei}$^{1,4}$\thanks{Corresponding Author.} \hspace{1.5em}
\textbf{Quanying Liu}$^{1,4,5}$\footnotemark[2] \\
\vspace{3mm} 
\centerline{$^1$Southern University of Science and Technology \quad $^2$University of Delaware} \\ 
\centerline{$^3$PSL Research University \quad $^4$Omni-Intelligence \quad $^5$Shenzhen Loop Area Institute} \vspace{1mm} 
\\
\centerline{\texttt{\url{https://github.com/ncclab-sustech/MindPilot}}}
\vspace{-6mm}
}

\begin{document}

\maketitle

\begin{abstract}
Whereas most brain–computer interface research has focused on decoding neural signals into behavior or intent, the reverse challenge—using controlled stimuli to steer brain activity—remains far less understood, particularly in the visual domain.
However, designing images that \textit{consistently elicit desired neural responses} is difficult: subjective states lack clear quantitative measures, and EEG feedback is both noisy and non-differentiable. 
We introduce \textbf{MindPilot}, a closed-loop framework that uses \textbf{EEG-derived feedback} to guide naturalistic image generation. MindPilot leverages non-invasive EEG with natural images, treats the brain as a black-box function, and employs a pseudo-model guidance mechanism to iteratively refine images without requiring access to gradients of the brain response. We evaluate MindPilot in simulations and human experiments, demonstrating (i) retrieval of semantic targets, (ii) closed-loop optimization of EEG features, and (iii) human-subject validation in an EEG-driven mental-matching task and a separate rating-driven emotion-regulation task. These results provide evidence for EEG-guided image synthesis in the evaluated settings and motivate further work on non-invasive closed-loop brain modulation, bidirectional brain--computer interfaces, and neural signal--guided generative modeling.
\end{abstract}

\section{Introduction}
   
The ability to modulate brain activity with precisely designed visual stimuli could open new avenues for cognitive enhancement, neurorehabilitation, and bidirectional human–AI interaction. However, designing images that reliably steer neural responses remains largely unexplored. Such modulation can be understood as steering the brain toward specific internal states, as reflected in neural signals like EEG~\citep{epstein1998cortical, qiu2023efficient}. 
Conceptually, this task resembles steering a deep visual encoder, where the aim is to design images that elicit particular internal representations. Yet, unlike artificial networks, the human brain poses unique challenges: humans' subjective states lack clear quantitative measures, real EEG responses are noisy and variable, and the brain itself is fundamentally \textit{non-differentiable}.

Recent advances in controllable generation, particularly text-conditioned diffusion models~\citep{li2019controllable, rahmani2022natural, epstein2023diffusion, ijcai2024p352}, offer unprecedented flexibility in image synthesis. But these models are optimized for linguistic prompts, not neural feedback, and thus remain orthogonal to the challenge of brain-targeted generation.
Prior efforts in closed-loop visual neuromodulation have shown that generative models can be guided by neuronal responses to synthesize activity-maximizing stimuli~\citep{ponce2019evolving, walker2019inception, bashivan2019neural, minai2024miso}. However, these approaches are typically invasive, relying on small-scale cortical recordings and targeting low-level neuronal activity rather than cognitive states. On the non-invasive side, EEG provides large-scale, distributed measures of brain activity that are more directly linked to cognition. For example,~\citep{luo2024vep} introduced the VEP Booster to modulate EEG with flickering visual stimuli. Yet such methods are constrained to low-level visual domains and cannot generate semantically rich, naturalistic images that engage higher-order brain representations.

\begin{figure}[t!]
\centering
\includegraphics[width=1\linewidth]{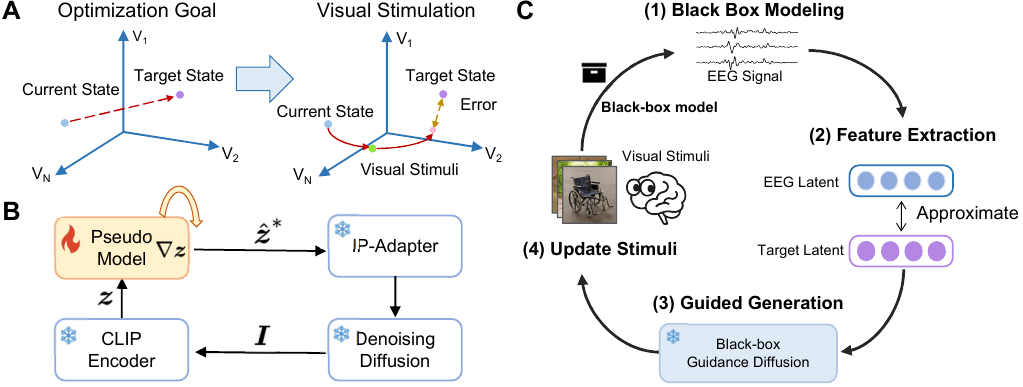}
\caption{\textbf{Conceptualization of MindPilot.} \textbf{A}. The goal of MindPilot is to continuously optimize visual stimuli to drive the brain latent state to the target. \textbf{B}. A pseudo-model provides surrogate gradients to iteratively refine images with respect to neural targets (e.g., semantic feature, spectral feature). \textbf{C}. The closed-loop visual optimization.}
\label{fig:overall}
\end{figure}

We introduce \textbf{MindPilot}, a closed-loop framework that unifies black-box optimization with diffusion-based generation for EEG-guided image design (Fig.~\ref{fig:overall}). MindPilot employs a \textit{surrogate-guided strategy} that replaces explicit reward gradients with pseudo-model updates, enabling gradient-free optimization toward diverse neural targets such as semantic similarity or EEG spectral features. Through both surrogate simulations and human EEG experiments, MindPilot achieves efficient \textit{convergence within limited iterations} while producing \textit{interpretable, naturalistic images aligned with neural selectivity and perception}. Our main contributions are:
\begin{itemize}
    \item \textbf{EEG-guided visual stimuli optimization:} a framework that uses black-box proxy feedback to guide diffusion-based synthesis toward EEG-derived targets.
    \item \textbf{Black-box guidance generation:} a pseudo-model strategy for gradient-free optimization of the neural targets evaluated in this work.
    \item \textbf{Multi-setting validation}: experiments with simulated EEG proxies and human participants, covering EEG semantic features, EEG spectral features, and self-reported emotional-valence ratings.
\end{itemize}

\begin{figure}[t]
\centering
\includegraphics[width=1.0\linewidth]{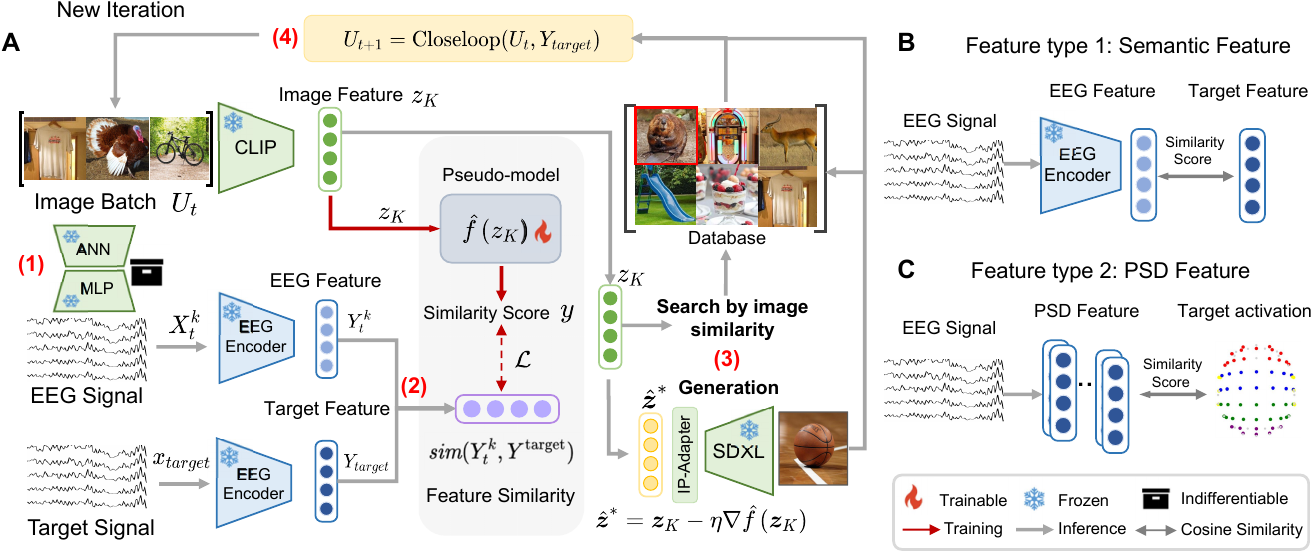}
\caption{\textbf{Framework}. \textbf{A}. Each MindPilot iteration involves four steps: 1): The black-box proxy model \(g\), which maps images to synthetic EEG, is designed as a black-box proxy model to predict the brain responses \(\bm{X}\). 2): The EEG Encoder \(f\) identifies different kinds of features \(\bm{Y}\) from EEG. MindPilot calculates the similarity score $sim(f(g(\bm{u})), \bm{y}_\text{target})$ as rewards based on EEG features. 3): Update the image embedding using gradient descent. 4): The image with a higher brain similarity score is selected and passed back to the image generator to optimize stimuli. \textbf{B}. Semantic feature from a pre-trained EEG encoder \(f\), aligned with CLIP embedding. \textbf{C}. Brain energy feature using Power Spectral Density (PSD) features. For more details, refer to Section~\ref{sec:method_framework}.} 
\label{fig:framework}
\end{figure}

\section{Related work}

\paragraph{Neural Selectivity and Invariance with EEG.}
EEG captures distributed neural signatures reflecting both \textit{selectivity} (e.g., the N170 ERP for faces~\citep{eimer2011face}) and \textit{invariance} (different inputs yielding equivalent neural responses~\citep{baroni2023learning}). Multivariate decoding methods have shown that EEG patterns can reliably differentiate object categories and semantic content~\citep{holm2024contribution}. However, most work treats EEG as an offline readout rather than \textit{an optimization signal for stimulus design}.

\paragraph{Closed-loop Visual Neuromodulation.}
Generative models have been used to design stimuli that maximize neural activity in invasive recordings~\citep{bashivan2019neural, walker2019inception, pierzchlewicz2023energy} and to refine low-level EEG biomarkers through flicker-based paradigms~\citep{luo2024vep}. 
In the semantic domain, prior non-invasive work has explored feedback-driven stimulus selection~\citep{grizou2025self} or generation based on binary attributes~\citep{Davis_2022_CVPR}. However, these approaches often operate on discrete choices or limited semantic subspaces, restricting the diversity of synthesized images. Similarly, methods relying on heuristic latent averaging~\citep{de2020brain} face challenges in sample efficiency during the search process.
MindPilot studies continuous naturalistic image synthesis driven by non-invasive EEG-derived feedback within this setting.

\paragraph{Brain-conditioned Controllable Generation.}
Text-conditioned diffusion has become the de facto standard for controllable image synthesis~\citep{nichol2022glide, ramesh2022hierarchical, ye2023ip, podell2024sdxl}. fMRI-conditioned image generation has been extensively developed~\citep{wang2024mindbridge,xia2024umbrae,scotti2024mindeye2,shen2024neuro,gong2025mindtuner}. Extensions to neural conditioning include gradient-based brain-guided synthesis~\citep{luo2023brain, luo2024brainscuba, cerdas2025brainactiv} and encoding-model alignment with fMRI~\citep{gu2023human,bao2025mindsimulator}. However, EEG-conditioned generation has mostly been explored for decoding or reconstruction~\citep{bai2024dreamdiffusion,li2024visual,fu2025brainvis,lopez2025guess,guo2025neuro}; closed-loop optimization of naturalistic image generation from EEG feedback remains comparatively underexplored. MindPilot studies this setting by optimizing generated images using EEG-derived responses.

\section{Methods}
\label{sec:method}
We introduce MindPilot, a closed-loop framework for generating the optimal visual stimuli to align with the target brain activity (Fig.~\ref{fig:framework}A). At its core, MindPilot treats the brain (or its EEG readout) as a \textit{black-box proxy model} and optimizes stimuli by iteratively maximizing the similarity between predicted neural responses and a target embedding. This formulation enables flexible objectives: e.g., aligning with \textit{semantic features} in CLIP space (Fig.~\ref{fig:framework}B) or \textit{EEG spectral features} (Fig.~\ref{fig:framework}C).

\subsection{Problem Setup}

Let $\bm{u}$ be a visual stimulus (an image) and let $\bm{x} \in \mathbb{R}^{C \times T}$ be the corresponding multi-channel EEG response it evokes, where $C$ is the number of channels and $T$ is the number of time points. We model the brain's visual processing as an unknown, non-differentiable forward process: $\bm{x} = g(\bm{u}),$
where the function $g$ can represent either the actual human brain or a pre-trained, black-box neural network serving as its proxy.

Our goal is to modulate the brain state towards a specific target. This target brain state is defined by a neural feature vector, $\bm{y}_\text{target}$, which is extracted from the specified target EEG signal $\bm{x}_\text{target}$ via a feature encoder, $f: \mathbb{R}^{C \times T} \to \mathbb{R}^{F}$. Thus, $\bm{y}_\text{target} = f(\bm{x}_\text{target})$. The objective is to find an optimal image, $\bm{u}^*$, that generates an EEG response whose features are maximally similar to the target features. This can be formulated as an optimization problem: $    \bm{u}^* = \arg\max_{\bm{u}} \; sim(f(g(\bm{u})), \bm{y}_\text{target}),$
where $sim(\cdot, \cdot)$ is the cosine similarity between the EEG representation evoked by visual stimuli $\bm{u}$ and the target brain state $\bm{y}_\text{target}$.

\subsection{Closed-loop Optimization}
\label{sec:method_framework}

MindPilot proceeds iteratively to find images matching a target feature ${\bm{y}}_\text{target}$. Starting from a uniform prior where every image has an equal selection probability $\frac{1}{N}$, MindPilot dynamically updates the score $\bm{S}_t(\bm{u})$ for each image in the database at iteration $t$. These scores define a non-uniform selection probability $P_t(\bm{u})$ that guides the sampling for the next round. Let $\mathcal{I}_{\text{best}}$ denote the set of indices corresponding to the top-$k$ images in step $t$ with the highest similarity scores (i.e., $|\mathcal{I}_{\text{best}}| = k$).

\textbf{Step A: Direct Reward Update.}
First, we compute an intermediate score $\bm{S}'_{t}$ by applying a direct reward to all images whose indices are in the best set $\mathcal{I}_{\text{best}}$. The score for each of these top-$k$ images is updated via an Exponential Moving Average (EMA):
\begin{equation}
\bm{S}'_{t}(\bm{u}_i) := 
\begin{cases} 
(1-\alpha) \cdot \bm{S}_{t}(\bm{u}_i) + \alpha \cdot sim( f(g(\bm{u}_i)), {\bm{y}}_\text{target} ) & \text{if } i \in \mathcal{I}_{\text{best}} \\
\bm{S}_{t}(\bm{u}_i) & \text{if } i \notin \mathcal{I}_{\text{best}},
\end{cases}
\label{eq:direct_update_topk}
\end{equation}
where $\alpha$ is the importance weight for the direct reward.

\textbf{Step B: Spreading Update.}
Next, we "spread" the rewards from the top-
$k$ images to other similar images in the database $\Omega = \{\bm{u}_1, \dots, \bm{u}_N\}$. The updated score for the $j$-th image, $\bm{S}_{t+1}$, is calculated by adding an aggregated spreading term, which is the average of the rewards spread from each of the top-$k$ images, weighted by their CLIP embedding similarity $s$:
\begin{equation}
\bm{S}_{t+1}(\bm{u}_j) := (1-\beta) \bm{S}'_{t}(\bm{u}_j) + \frac{\beta}{|\mathcal{I}_{\text{best}}|} \sum_{i \in \mathcal{I}_{\text{best}}} \bm{S}'_{t}(\bm{u}_i) \frac{\exp(s(\bm{u}_i, \bm{u}_j))}{\sum_{l=1}^{N}\exp(s(\bm{u}_i, \bm{u}_l))},
\label{eq:spread_update_topk}
\end{equation}
where $\beta$ is the hyperparameter. This ensures that an image $\bm{u}_i$ receives a stronger score boost if it is similar to multiple images in the high-scoring set $\mathcal{I}_{\text{best}}$. The reward being spread from each best item is proportional to its own updated score, $\bm{S}'_{t}(\bm{u}_i)$.

Having defined a valid score $\bm{S}_{t+1}(\bm{u}_j)$ for all candidate images $j \in \{1, \dots, N\}$ with Eq. (1) and 2, the updated scores are converted into a probability distribution for the next selection round using the softmax function:
\begin{equation}
P_{t+1}(\bm{u}_j) := \frac{\exp(\bm{S}_{t+1}(\bm{u}_j))}{\sum_{l=1}^{N} \exp(\bm{S}_{t+1}(\bm{u}_l))}.
\label{eq:prob_update_final}
\end{equation}
The framework then samples images based on this updated probability $P_{t+1}(\bm{u}_j)$ to continue the search and generation process in the subsequent iteration $t+1$.

\subsection{Black-box Proxy Model}
\label{sec:method_encoding_model}
Direct real-time EEG acquisition is costly and impractical for large-scale closed-loop experiments. To address this, MindPilot is designed to interface flexibly with an EEG black-box proxy model $g$, serving as a black-box proxy of the human brain, which is treated as a non-differentiable black-box in optimization. The proxy model combines a pre-trained backbone with a regression head trained to minimize Mean Square Error (MSE) against real EEG recordings. Specifically, we replace the classification layer of the backbone with a \(C \times T\) regression layer, where each unit corresponds to a flattened EEG channel–timepoint signal. 

To demonstrate the framework’s flexibility, we instantiated a diverse set of black-box proxies, ranging from classic CNNs (AlexNet~\citep{krizhevsky2012imagenet}, ResNet50~\citep{he2016deep}, CORnet-S~\citep{kubilius2019brain}) to recent self-supervised and vision-transformer architectures (MoCo~\citep{he2020momentum}, ViT-B-32~\citep{dosovitskiy2021an}, OpenCLIP-ViT-B-32~\citep{cherti2023reproducible}, DINO2-ViT-B-14~\citep{oquab2024dinov}, DINO-ViT-B-16~\citep{caron2021emerging}, SYNCLR-ViT-B-16~\citep{sundaram2024does}), as reported in Tab.~\ref{tab:compare_readout_model}.

\begin{table}[ht]
\centering
\caption{\textbf{Evaluation of black-box proxy models.} 
We compared the predicted visual-evoked EEG response against the real EEG response over the [60, 500] ms post-stimulus window. 
We report the \textbf{averaged} Pearson's correlation coefficient $R$ and the lower Noise Ceiling as group mean (sample standard deviation) across 10 subjects. 
\textit{Note:} This metric aggregates performance across a broad window, which inherently dilutes peak correlations. Time-resolved analysis reveals significantly stronger predictive power at key latencies (reaching $r \approx 0.6$ at $\sim$100ms, see \textbf{Appendix Fig.~\ref{fig:pearson_timepoint}}).}
\setlength{\tabcolsep}{3pt}
\resizebox{\textwidth}{!}{%
\begin{tabular}{lccccccccc|c}
\toprule
Metrics(\%) & AlexNet & ResNet50 & CORnet-S & MoCo & ViT & OpenCLIP & DINO & DINO2 & SYNCLR & Lower Noise Ceiling \\
\midrule
\multicolumn{1}{c}{$R \uparrow$} & $33.09\,(3.93)$ & $33.49\,(4.10)$ & \underline{$33.56\,(4.25)$} & $32.68\,(3.95)$ & $31.27\,(4.32)$ & $33.40\,(4.36)$ & $33.39\,(4.20)$ & $\bm{33.70\,(4.64)}$ & $27.69\,(3.89)$ & $38.75\,(6.26)$ \\
\bottomrule
\end{tabular}%
}
\label{tab:compare_readout_model}
\end{table}

\subsection{Interactive Search}
\label{sec:method_retrieval}

To bootstrap optimization with an unknown target image, drawing inspiration from interactive retrieval methods~\citep{ferecatu2007interactive}, we propose a similarity-weighted sampling approach. Starting from random candidates, MindPilot updates the sampling distribution using a roulette-wheel selection weighted by similarity score. Over iterations, the distribution sharpens around stimuli eliciting neural activity closer to the target (Algorithm~\ref{alg:interactive_search}).

The process begins with a randomly selected set of images $U_0$, without prior knowledge of the specific features of the target image. We use the roulette wheel selection algorithm to choose from current images based on the similarity score $sim(f(g(\bm{u}^k_+)), \bm{y}_{\text{target}})$. The system updates the probability $\bm{P}_t(\bm{u}^k_+)$ for each image in the database belonging to the target class, based on the response model's prediction \(\bm{Y}_t = f(g(\bm{U}_t)) \in \mathbb{R}^{N \times F}\). Subsequently, MindPilot calculates the similarity score between the target and the feature predicted by the image selected. See more implementation details in Appendix~\ref{sec:Retrieval_pipeline}. Once an image is identified as the best in a iteration, the likelihood of similar images in the search space belonging to the target class is increased.

\subsection{Heuristic Generation}
\label{sec:method_generation}
Searching in a fixed image pool is limiting. We therefore integrate EEG-guided diffusion to generate new stimuli. Specifically, we employ the pretrained guided diffusion model SDXL-Lightning~\citep{lin2024sdxl_lightning} with black-box guidance: a Gaussian Process (GP) surrogate predicts reward gradients in latent space for EEG-guided image generation.  

\paragraph{EEG-driven Black-box Guidance}
Recently, black-box guidance diffusion has proved to be successful at generating high-quality images and drug discovery~\citep{fan2023dpok, black2024training, tan2025fast}. To replace directly calculating the gradient on the black-box encoding model, we use a pseudo target embedding $\hat{z}^*$ by the pseudo target model with input $z_K$. For the Gaussian Process (GP) update case, we set the pseudo target embedding $\hat{z}^*$ as the one gradient step update using the gradient of the mean prediction of the GP surrogate model as follows:
\begin{equation}
{\hat{\bm{z}}^*} = \bm{z}_K - \eta  \nabla \hat{f}(\bm{z}_K; \bm{Z}^n), 
\end{equation}
where $K$ is the number of diffusion sampling steps, $\eta$ is the step size that decays linearly, $\bm{z}_K$ denotes the CLIP embedding from generated latents and $\bm{Z}^n = [\bm{z}^1, \cdots, \bm{z}^n]$. The $\hat{f}(\bm{z}_K; \bm{Z}^n)$ denotes the prediction of the GP surrogate model~\citep{seeger2004gaussian} evaluated at $z_K$, which has closed-form as below:
\begin{equation}
\hat{f}(\bm{z}_K; \bm{Z}^n) = \bm{k}(\bm{z}_K, \bm{Z}^n)^T \left( \mathcal{K}(\bm{Z}^n, \bm{Z}^n) + \lambda \bm{I} \right)^{-1} y
\end{equation}
where $\bm{Z}^n = [\bm{z}^1, \cdots, \bm{z}^n]$ and $\bm{y} = [y^1, \cdots, y^n]$ denotes the collected data from the previous $t-1$ step and its corresponding rewards in each sampled set $U_t$, respectively. We define each reward $y^i$ by the scaling factor $\gamma$ as $y^i = sim( f(g(\bm{u}_i)), \bm{y}_\text{target} ) \times \gamma$.

\paragraph{Online Iteration Algorithm}
We integrate an EEG-guided genetic algorithm to optimize stimuli towards the evolution direction of the target neural activity. The specific procedural steps of our algorithm are outlined in Algorithm~\ref{alg:heuristic_generation}. Unlike the interactive searching process in Algorithm~\ref{alg:interactive_search}, after sampling the stimulus image in each step $t$, we perform crossover and "mutation" on the image embeddings and then sample new images from the image space. See Appendix~\ref{sec:Generation_pipeline} for additional details on the evolution process. Throughout this process, the relative order of the original CLIP features is preserved within each dimension to ensure that the mutated images remain semantically coherent and interpretable by humans.

\begin{figure}[ht]
\centering
\caption{\textbf{EEG-guided Interactive Search.} \textbf{A}. Similarity score between MindPilot’s neural representations (steps 1, 2, last, and best-step) and the target, versus random stimuli. \textbf{B}. Cross-modal correlation between image and EEG embedding similarity ($R=0.23$, $P<0.01$). \textbf{C}. Similarity score improvement across all subjects using semantic features. \textbf{D}. The correlation between image embedding similarity and EEG semantic feature similarity across all subjects. The vertical axis represents the similarity score between the EEG features at the current step and the target.}
\includegraphics[width=1.0\linewidth]{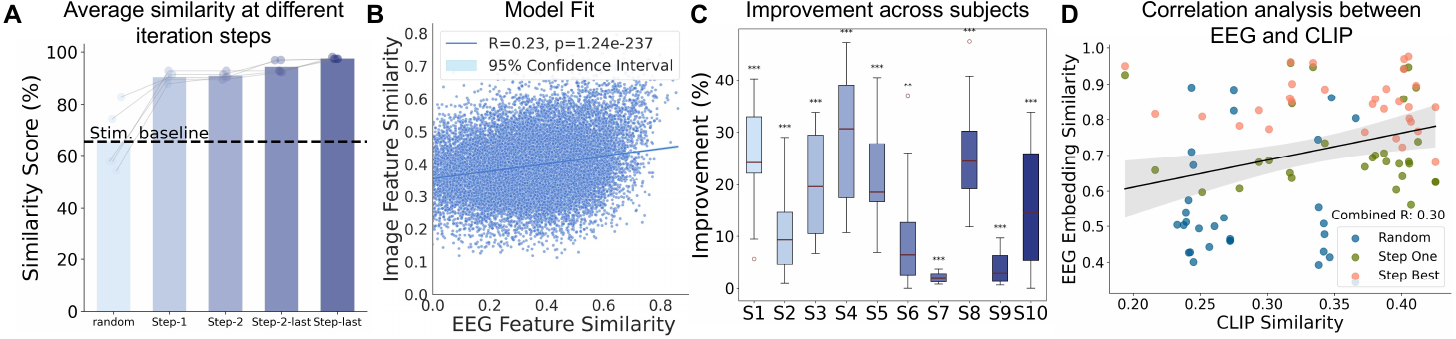}
\label{fig:retrieval}
\end{figure}

\section{Experiments}
\label{sec:experiments}

\subsection{Setup}
\paragraph{EEG readout from proxy models} We trained a family of predictors to approximate EEG responses evoked by visual stimuli using the THINGS-EEG2 dataset~\citep{gifford2022large, grootswagers2022human}. Each model regressed 17-channel $\times$ 250 timepoint EEG signals from pre-trained visual features. Nine variants were evaluated across participants, with performance quantified by Pearson’s $r$ and noise ceiling (Tab.~\ref{tab:compare_readout_model}). Training used all repetitions (four trials per image) and was computationally efficient, fitting within 48 GB memory on a single NVIDIA L40 GPU (see Appendix~\ref{sec:appendix_datasets} for details).
As shown in Tab.~\ref{tab:compare_readout_model}, relatively simple CNN-based proxies (e.g., AlexNet and ResNet50) also achieved competitive prediction accuracy. Within the evaluated architectures, MindPilot was not tied to a single proxy backbone and could be used with multiple image-to-EEG predictors.

\paragraph{Target Features of the evoked EEG} 
\label{target}
We considered two neural targets: (i) semantic embeddings, extracted from ATM-S~\citep{li2024visual} and aligned with CLIP ViT-H-14 features, and (ii) spectral signatures, derived from EEG power spectral density (PSD). Semantic targets were used in the retrieval task, while both semantic and PSD targets were used in the generation task, ensuring that optimization was cognitively meaningful. Moreover, to test whether MindPilot framework can be extended to the subjective state modulation even without an explicit EEG feature as the target, we considered emotion regulation under visual stimulation using self-reported scores in real human experiment in Appendix~\ref{sec:human_exp_config_results_emotion}.

\subsection{EEG-guided Interactive Search}

We first tested whether MindPilot can retrieve a target stimulus from the THINGS-EEG2 search space (50 categories × 12 conditions = 600 images). Starting from 10 random images, the model iteratively updated a sampling distribution based on EEG–CLIP similarity. We set slippage factor ($\alpha=0.1$) and reward propagation factor ($\beta=0.1$). Although the set hyperparameters were not subjected to a thorough search, they were sufficient to bring about an increase in iterations. Results are summarized in Fig.~\ref{fig:retrieval}. \textbf{Convergence}: Similarity scores improved consistently across iterations, surpassing random sampling baselines (Fig.~\ref{fig:retrieval}A).
\textbf{Alignment}: EEG embeddings exhibited significant correlations with CLIP representations across participants, confirming CLIP similarity as a valid proxy for neural alignment (Fig.~\ref{fig:retrieval}B).
\textbf{Subject-level consistency}: Per-subject performance (averaged across five seeds and three target images) showed improvements, confirmed by paired $t$-tests (Fig.~\ref{fig:retrieval}C).
\textbf{Dynamics}: Iterative updates progressively aligned EEG and CLIP embeddings (Fig.~\ref{fig:retrieval}D).
These results support MindPilot's closed-loop retrieval behavior under the evaluated EEG-proxy setting.

\begin{figure}[t!]
\centering
\includegraphics[width=1\textwidth]{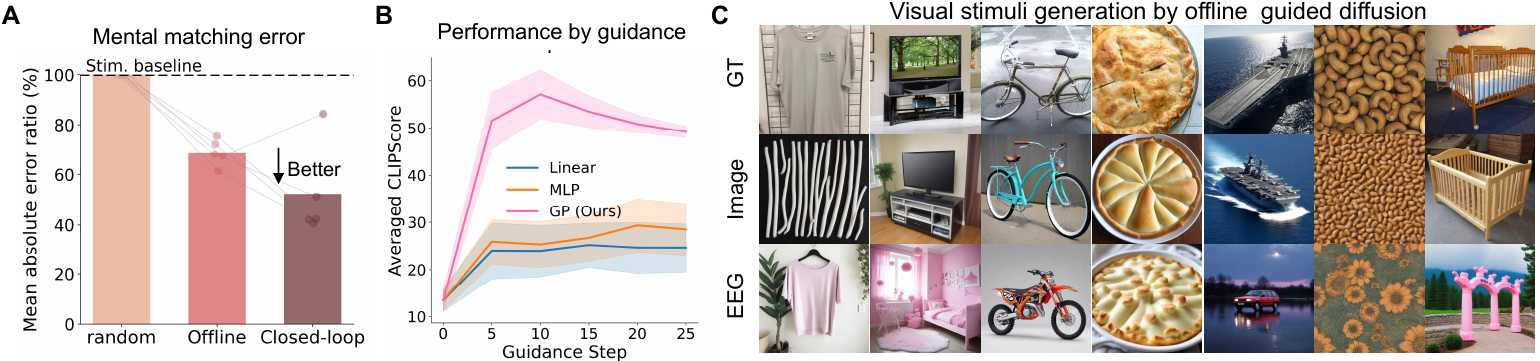}
\caption{\textbf{EEG semantic feature-guided visual stimuli generation.} 
\textbf{A}. Mean L1 error reduction relative to the random stimuli baseline and offline EEG-guided generation, averaged over five closed-loop experimental targets.
\textbf{B}. CLIP similarity scores for semantic reconstruction across three different estimators in the pseudo-model setup.
\textbf{C}. Ablation study on optimization guidance modalities. Top: Original ground-truth target images. Middle (Oracle reference): generation results optimized directly using target-image CLIP features. Bottom (Ours): generation results optimized using EEG features (MindPilot).} 
\label{fig:generation_semantic}
\end{figure}

\subsection{EEG-guided Stimulus Generation}

We next evaluated whether MindPilot can generate optimized stimuli rather than search in a fixed pool. Four candidates were selected for the next round via roulette sampling. We conducted a closed-loop stimulus optimization experiment within the 600-image space (50 categories × 12 conditions) of THINGS-EEG2. Starting from 10 randomly selected images (out of 200), we used Stable Diffusion XL-Lightning~\citep{lin2024sdxl_lightning} with IP-Adapter~\citep{ye2023ip} to generate 2 new images per iteration. From the resulting image feature space, 4 candidates were selected via roulette sampling for the next round of stimulation. Tab.~\ref{tab:add_iteration_performance} reports different feature-based Semantic Similarity Score (SS) and Intensity Similarity Score (IS) scores, showing that our method effectively identifies target stimulus within limited iterations. Additional quantitative results are in Appendix~\ref{sec:add_experimental_results}. 
These results support the efficiency of MindPilot across the brain-modulation targets evaluated here.

\begin{table}[h]
\centering
\caption{\textbf{Performance of MindPilot in EEG semantic-driven image generation.} We compare MindPilot with two EEG-to-image methods (i.e., ATM-S~\citep{li2024visual} and CognitionCapturer~\citep{zhang2025cognitioncapturer}); ATM-S is reported as an oracle/reference ceiling obtained under a different, direct-reconstruction protocol. The chance-level baseline for Subject-01 is also shown. Bold indicates the best performance (optimal model). Underlined indicates the second-best performance (suboptimal model).}
\setlength{\tabcolsep}{4pt}
\label{tab:reconstruction}
\resizebox{\textwidth}{!}{%
\begin{tabular}{llccccccc}
\toprule
\multirow{3}{*}{Type} & \multirow{3}{*}{Method} & \multicolumn{2}{c}{Low-level}     &     \multicolumn{5}{c}{High-level}     \\
\cmidrule(r){3-4} \cmidrule(l){5-9}  
&& PixCorr $\uparrow$ & SSIM $\uparrow$ & AlexNet(2) $\uparrow$ & AlexNet(5) $\uparrow$ & Inception $\uparrow$ & CLIP $\uparrow$ & SwAV $\downarrow$  \\
\midrule
\multirow{2}{*}{EEG-to-image} & ATM-S (Oracle reference) & 0.14 & 0.32 & \textbf{0.80} & \textbf{0.85} & \textbf{0.72} &   \textbf{0.76} & \textbf{0.58} \\
& CongCapturer & \textbf{0.15} & \textbf{0.33}  & 0.73  & 0.81 & 0.65 & 0.68 & 0.59 \\
\midrule
\multirow{2}{*}{Modulation} & Chance-level & 0.05 & 0.28 & 0.49  & 0.49 & 0.50 & 0.48 & 0.69 \\
& \cellcolor{mylightgray}MindPilot (Ours) & \cellcolor{mylightgray}\textbf{0.09} & \cellcolor{mylightgray}\textbf{0.35} & \cellcolor{mylightgray}\textbf{0.70} & \cellcolor{mylightgray}\textbf{0.73} & \cellcolor{mylightgray}\textbf{0.58} & \cellcolor{mylightgray}\textbf{0.67} & \cellcolor{mylightgray}\textbf{0.60} \\
\bottomrule
\end{tabular}%
}
\end{table}

\paragraph{Semantic-driven generation} In a 10-round EEG semantic-guided visual stimuli generation experiment, MindPilot was compared against random selection and an offline black-box diffusion baseline. Across five target images, closed-loop optimization consistently reduced L1 error relative to baselines (Fig.~\ref{fig:generation_semantic}A). Generated images preserved semantic fidelity, as illustrated by comparisons with offline GP-based estimators (Fig.~\ref{fig:generation_semantic}B–C). In Fig.~\ref{fig:generation_semantic}, the middle row is an oracle reference obtained with noiseless target-image features. Tab.~\ref{tab:reconstruction} also reports specialized EEG-to-image decoders (ATM-S~\citep{li2024visual} and CognitionCapturer~\citep{zhang2025cognitioncapturer}). Because these decoders map ground-truth EEG directly to images under a different protocol, their results are used as reference ceilings rather than strict theoretical upper bounds for MindPilot. MindPilot outperforms the chance-level baselines (e.g., SSIM: 0.35 vs. 0.28) and approaches the high-level semantic alignment of the reference method in the reported comparison (e.g., CLIP-2 way: 0.67 vs. 0.76 for ATM-S).

These results show that, in the evaluated setting, MindPilot's closed-loop optimization steers image generation toward neural targets with semantic fidelity approaching that of the reported specialized decoders, despite the different task protocols.

\begin{table}[htbp]
  \centering
  \caption{\textbf{Performance of closed-loop iteration using semantic embedding}. We used the pretrained AlexNet as the backbone of the black-box proxy model. We evaluate the performance using two metrics: Semantic Similarity Score (SS) and Intensity Similarity Score (IS). We reported per-subject SS and IS values, from initial stimuli (Random) and from optimized stimuli (Step-1 and Step-Best). Ratios of improvements by the closed-loop framework (i.e. $\Delta$SS, $\Delta$IS) are reported.}
  \label{tab:add_iteration_performance}
  
  \begingroup 
  \setlength{\tabcolsep}{4pt} 
  \renewcommand{\arraystretch}{1.0} 
  \footnotesize 
  
  \begin{tabular}{c | cc | cc | cc | cc} 
    \hline
      & \multicolumn{2}{c|}{Random}
      & \multicolumn{2}{c|}{Step-1}
      & \multicolumn{2}{c|}{Step-Best}
      & \multicolumn{2}{c}{Improved by MindPilot} \\
    \cmidrule(r){2-3} \cmidrule(r){4-5} \cmidrule(r){6-7} \cmidrule(r){8-9}
    Subject & SS & IS & SS & IS & SS & IS & $\Delta$SS (\%) & $\Delta$IS (\%) \\
    \hline
    1  & 0.5174 & 0.9632 & 0.6686 & 0.9729 & \textbf{0.8375} & \textbf{0.9976} & 16.8859 & 2.4790 \\
    2  & 0.5197 & 0.9678 & 0.6675 & 0.9764 & \textbf{0.7372} & \textbf{0.9998} & 6.9701  & 2.3406 \\
    3  & 0.5113 & 0.9883 & 0.6597 & 0.9927 & \textbf{0.7871} & \textbf{0.9980} & 12.7402 & 0.5306 \\
    4  & 0.5065 & 0.9650 & 0.6498 & 0.9836 & \textbf{0.8299} & \textbf{0.9963} & 18.0136 & 1.2690 \\
    5  & 0.5315 & 0.9788 & 0.6937 & 0.9768 & \textbf{0.8418} & \textbf{0.9979} & 14.8151 & 2.1055 \\
    6  & 0.6747 & 0.9836 & 0.8099 & 0.9856 & \textbf{0.8826} & \textbf{0.9961} & 7.2634  & 1.0461 \\
    7  & 0.8838 & 0.8955 & 0.9410 & 0.9033 & \textbf{0.9500} & \textbf{0.9742} & 1.8237  & 7.0879 \\
    8  & 0.5077 & 0.8344 & 0.6838 & 0.9435 & \textbf{0.8568} & \textbf{0.9925} & 17.3066 & 4.8947 \\
    9  & 0.8465 & 0.9602 & 0.9251 & 0.9751 & \textbf{0.9597} & \textbf{0.9997} & 3.4662  & 2.4597 \\
    10 & 0.5128 & 0.8172 & 0.6707 & 0.9705 & \textbf{0.7687} & \textbf{0.9934} & 9.8032  & 2.2849 \\
    \hline
    \rowcolor{mylightgray} Average
       & 0.6012 & 0.9354
       & 0.7370 & 0.9680
       & \textbf{0.8451} & \textbf{0.9946}
       & 10.9088 & 2.6498 \\
    \hline
  \end{tabular}
  \endgroup
\end{table}

\begin{figure}[t!]
\centering
\includegraphics[width=1\textwidth]{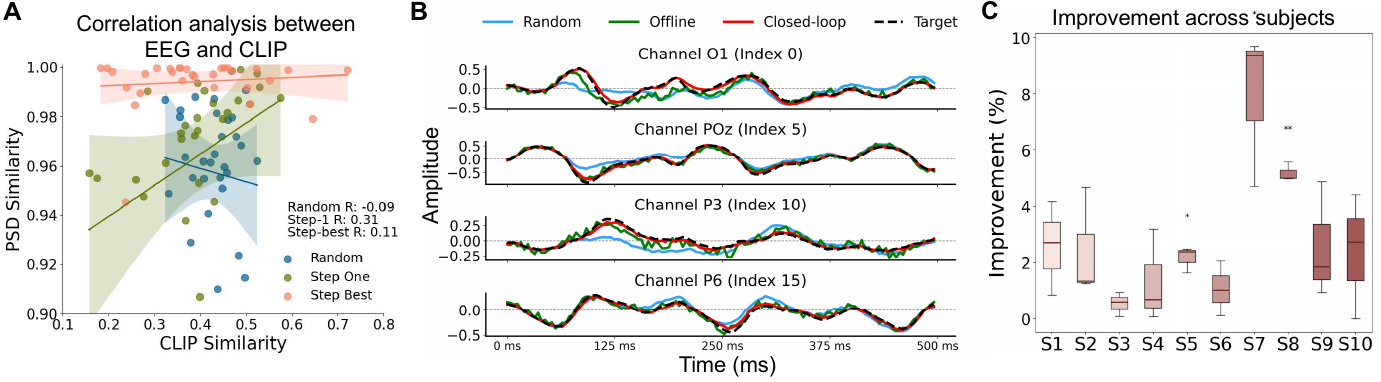}
\caption{\textbf{EEG PSD-driven generation.} \textbf{A}. Neural-visual correlation analysis. We report the correlation coefficients between EEG PSD feature similarity and image CLIP feature similarity across 10 subjects in the THINGS-EEG2 dataset. \textbf{B}. Examples of heuristic generation guided by PSD feature across EEG channels. \textbf{C}. Improvement in feature similarity scores across all 10 subjects, with statistical significance determined by paired t-tests. } 
\label{fig:generation_psd}
\end{figure}

\paragraph{PSD-driven generation} MindPilot also optimized toward PSD features, directly modulating neural spectral patterns. Correlation analysis confirmed progressive alignment between image features and EEG PSD (Fig.~\ref{fig:generation_psd}A). The comparisons across the random baseline, initial step (step-1), and optimized step (step-best) demonstrate that our iteration effectively aligns neural spectral patterns with visual semantics. Neural alignment was particularly evident in the early post-stimulus window (0–500 ms), where closed-loop images evoked EEG responses closely matching the target (Fig.~\ref{fig:generation_psd}B). Across subjects and seeds, PSD-guided generation significantly improved similarity scores (Fig.~\ref{fig:generation_psd}C). 

Together, these findings demonstrate that MindPilot extends beyond search to actively design stimuli that align with both semantic and spectral targets.

\begin{figure}[ht]
\centering
\includegraphics[width=1\textwidth]{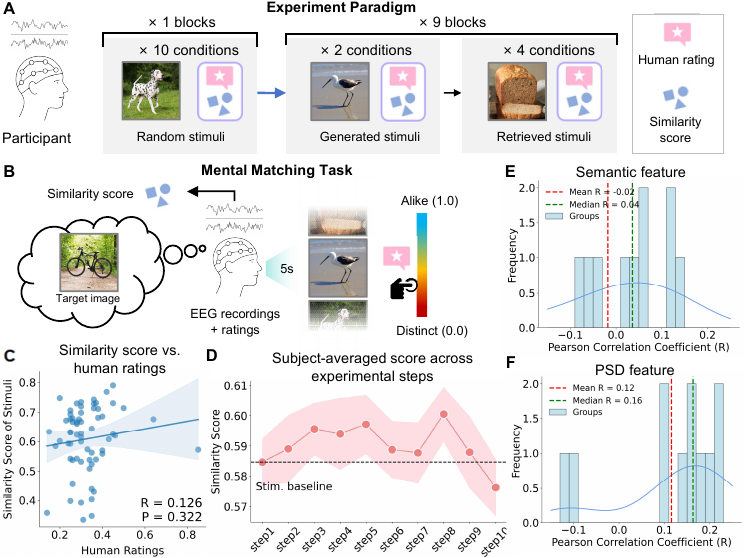}
\caption{\textbf{Real-time EEG-driven mental-matching experiment.} \textbf{A}. Overview of the experimental paradigm, involving sequential image presentation and rating blocks. \textbf{B}. Target-guided similarity judgment: participants (N=10) first memorized a target image, then rated subsequent images on their similarity score to the target, ranging from 0 [least similar] to 1 [highest similar]. \textbf{C}. Example correlation between model-derived similarity scores and human ratings for a single target from one participant (Participant, Target 1). \textbf{D}. The mean and std. of similarity score from step-1 to step-10 stimuli generated by MindPilot across all 10 participants. \textbf{E-F}. Correlation between the similarity score and human ratings.}
\label{fig:human_exps}
\end{figure}

\subsection{Real-time Closed-loop Human Experiments}
To evaluate MindPilot beyond proxy models, we recruited 10 participants under a protocol approved by the Science and Technology Ethics Committee of Southern University of Science and Technology. All participants provided written informed consent. Group-level trajectory analyses used all 10 participants. Four participants were excluded from the correlation analyses because their recordings were classified as having poor data quality, leaving 6 participants for those analyses; the available study materials do not document a quantitative exclusion threshold. Appendix~\ref{sec:add_human_exp} reports the protocol and analysis subsets.
In the mental-matching task, participants memorized a target image and then rated optimized images on a [0,1] similarity scale (Fig.~\ref{fig:human_exps}A,B). EEG recorded during image viewing supplied the real-time feedback used for adaptive stimulus optimization; the ratings were used to evaluate perceived similarity to the memorized target.

Results are shown in Fig.~\ref{fig:human_exps}C--F. Model-derived similarity scores from EEG responses correlated with human judgments (Fig.~\ref{fig:human_exps}C), and participant ratings increased across iterations (Fig.~\ref{fig:human_exps}D). The mental-matching analyses in Fig.~\ref{fig:human_exps}E--F use the six-participant correlation subset described above. The separate emotion-regulation task and its rating-driven feedback are reported in Fig.~\ref{fig:human_exp_config_results_emotion}.

\subsubsection{Human Emotion Regulation Experiments}
\label{sec:human_exp_config_results_emotion}
\begin{figure}[htbp]
    \centering
    \includegraphics[width=1.0\linewidth]{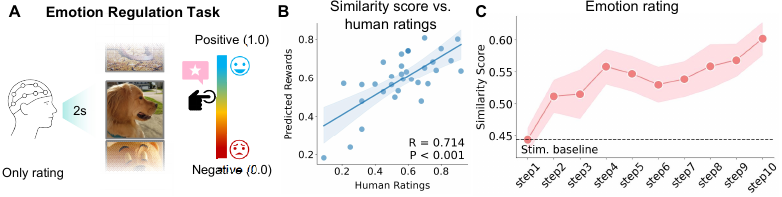}
    \caption{\textbf{A}. Affective evaluation: participants rated along an affective dimension, ranging from 0 [sad] to 1 [happy].  \textbf{B}. Correlation between the predicted rewards from the estimator and human ratings in the emotion regulation task (i.e., Participant 4, Target 1). \textbf{C}. The mean and std. of similarity score from step-1 to step-10 stimuli generated by MindPilot across all 10 participants.}
    \label{fig:human_exp_config_results_emotion}
\end{figure}

\textbf{Emotion regulation experiment}. Participants rated each image based on the evoked emotional valence, using a continuous scale from 0 (negative: sad, unpleasant) to 1 (positive: happy, pleasant). Unlike mental matching, this task used the participant's subjective rating, not EEG, as the feedback signal for closed-loop optimization. The control goal was to shift self-reported valence toward the positive end of the scale. Fig.~\ref{fig:human_exp_config_results_emotion}A diagrams the rating-driven experiment. Fig.~\ref{fig:human_exp_config_results_emotion}B reports the association between pseudo-model estimates and participant ratings (R = 0.714, P $\leq$ 0.001), and Fig.~\ref{fig:human_exp_config_results_emotion}C reports the group-level trajectory (0.45$\rightarrow$0.60).

\section{Discussion and conclusion}
\label{sec:Discussion}
We presented \textbf{MindPilot}, a closed-loop framework for EEG-guided visual optimization with naturalistic images. Across the evaluated proxy readouts, simulation-based optimization, and human mental-matching experiment, MindPilot moved generated stimuli toward the specified neural targets while preserving interpretable image content. The separate emotion-regulation experiment showed that the same optimization structure can also operate with subjective ratings as feedback. Together, these experiments provide a proof of concept for black-box brain--stimulus co-optimization under the protocols studied here.

\paragraph{Technical Impact}
MindPilot introduces a surrogate-guided optimization strategy that enables gradient-free guidance of diffusion models toward EEG-derived objectives. In the reported experiments, this design preserved image quality while optimizing toward EEG-derived representations; the rating-driven experiment separately optimized self-reported affective valence. These results motivate further investigation in brain--computer interfaces, cognitive neuroscience, and non-invasive neuromodulation~\citep{jang2021noise, alamia2023role}.

\paragraph{Limitations and future directions}
While our results are promising, several challenges remain.
First, multiple distinct stimuli can elicit similar EEG responses, echoing Metamers~\citep{feather2023model}. Future work should disentangle the representational factors underlying such degeneracy of EEG-conditioned generation.
Second, current models do not explicitly address inter-subject variability. Incorporating subject-adaptive or transfer learning strategies could improve cross-participant consistency and individual control accuracy~\citep{alamia2023role}. Moreover, although the mental-matching study used real-time EEG feedback, the system was not optimized for low latency. Lower-latency continuous operation remains an important direction.
Lastly, it is important to extend MindPilot to other modalities to provide richer neural feedback and support higher-level control objectives (e.g., attention, memory, or affective modulation).

\clearpage
\section*{Ethics Statement}
All human-subject experiments were conducted in accordance with the Declaration of Helsinki. The protocol was reviewed and approved by the Science and Technology Ethics Committee of Southern University of Science and Technology (SUSTech; decision no.~20250292, acceptance no.~2025PES292) on June 30, 2025, under the project ``Emotion-related EEG decoding, modeling, and neurofeedback regulation in naturalistic settings,'' an English translation of the project title in the approval document. All 10 participants were informed of the study procedures and goals and provided written informed consent. Participation was voluntary, participants could withdraw at any time without penalty, and compensation was based on participation duration. The study involved non-invasive EEG recording, visual stimulation, mental-matching ratings, and emotional-valence ratings. We also used the publicly available THINGS-EEG2 dataset, which was collected and shared under its own ethical framework.

\section*{Acknowledgement}
This work was supported by the National Natural Science Foundation of China (62472206), National Key R\&D Program of China (2025YFC3410000), Shenzhen Science and Technology Innovation Committee (RCYX20231211090405003, JCYJ20220818100213029), GuangDong Basic and Applied Basic Research Foundation (2025A1515011645 to ZC.L.), Shenzhen Doctoral Startup Project (RCBS20231211090748082 to XK.S.), Guangdong Provincial Key Laboratory of Advanced Biomaterials (2022B1212010003), and the open research fund of the Guangdong Provincial Key Laboratory of Mathematical and Neural Dynamical Systems, the Center for Computational Science and Engineering at Southern University of Science and Technology.

\bibliography{iclr2026_conference}

@article{bashivan2019neural,
  title={Neural population control via deep image synthesis},
  author={Bashivan, Pouya and Kar, Kohitij and DiCarlo, James J},
  journal={Science},
  volume={364},
  number={6439},
  pages={eaav9436},
  year={2019},
  publisher={American Association for the Advancement of Science}
}

@article{ponce2019evolving,
  title={Evolving images for visual neurons using a deep generative network reveals coding principles and neuronal preferences},
  author={Ponce, Carlos R and Xiao, Will and Schade, Peter F and Hartmann, Till S and Kreiman, Gabriel and Livingstone, Margaret S},
  journal={Cell},
  volume={177},
  number={4},
  pages={999--1009},
  year={2019},
  publisher={Elsevier}
}

@article{epstein2023diffusion,
  title={Diffusion self-guidance for controllable image generation},
  author={Epstein, Dave and Jabri, Allan and Poole, Ben and Efros, Alexei and Holynski, Aleksander},
  journal={Advances in Neural Information Processing Systems},
  volume={36},
  pages={16222--16239},
  year={2023}
}

@article{luo2024vep,
  title={The VEP Booster: A Closed-Loop AI System for Visual EEG Biomarker Auto-generation},
  author={Luo, Junwen and Jiang, Chengyong and Chen, Qingyuan and Han, Dongqi and Wang, Yansen and Yan, Biao and Li, Dongsheng and Zhang, Jiayi},
  journal={arXiv preprint arXiv:2407.15167},
  year={2024}
}

@inproceedings{
    luo2024brainscuba,
    title={Brain{SCUBA}: Fine-Grained Natural Language Captions of Visual Cortex Selectivity},
    author={Andrew Luo and Margaret Marie Henderson and Michael J. Tarr and Leila Wehbe},
    booktitle={The Twelfth International Conference on Learning Representations},
    year={2024},
    url={https://openreview.net/forum?id=mQYHXUUTkU}
}

@InProceedings{Davis_2022_CVPR,
    author    = {Davis, III, Keith M. and de la Torre-Ortiz, Carlos and Ruotsalo, Tuukka},
    title     = {Brain-Supervised Image Editing},
    booktitle = {Proceedings of the IEEE/CVF Conference on Computer Vision and Pattern Recognition (CVPR)},
    month     = {June},
    year      = {2022},
    pages     = {18480-18489}
}

@article{wang2023better,
  title={Better models of human high-level visual cortex emerge from natural language supervision with a large and diverse dataset},
  author={Wang, Aria Y and Kay, Kendrick and Naselaris, Thomas and Tarr, Michael J and Wehbe, Leila},
  journal={Nature Machine Intelligence},
  volume={5},
  number={12},
  pages={1415--1426},
  year={2023},
  publisher={Nature Publishing Group UK London}
}

@inproceedings{song2024decoding,
    title={Decoding Natural Images from {EEG} for Object Recognition},
    author={Yonghao Song and Bingchuan Liu and Xiang Li and Nanlin Shi and Yijun Wang and Xiaorong Gao},
    booktitle={The Twelfth International Conference on Learning Representations},
    year={2024},
    url={https://openreview.net/forum?id=dhLIno8FmH}
}

@inproceedings{geman2006interactive,
  title={Interactive image retrieval by mental matching},
  author={Geman, Donald},
  booktitle={Proceedings of the 8th ACM international workshop on Multimedia information retrieval},
  pages={1--2},
  year={2006}
}

@inproceedings{gong2025mindtuner,
  title={MindTuner: Cross-Subject Visual Decoding with Visual Fingerprint and Semantic Correction},
  author={Gong, Zixuan and Zhang, Qi and Bao, Guangyin and Zhu, Lei and Xu, Rongtao and Liu, Ke and Hu, Liang and Miao, Duoqian},
  booktitle={Proceedings of the AAAI Conference on Artificial Intelligence},
  volume={39},
  pages={14247--14255},
  year={2025},
  doi={10.1609/aaai.v39i13.33560}
}

@inproceedings{cerdas2025brainactiv,
title={Brain{ACTIV}: Identifying visuo-semantic properties driving cortical selectivity using diffusion-based image manipulation},
author={Diego Garcia Cerdas and Christina Sartzetaki and Magnus Petersen and Gemma Roig and Pascal Mettes and Iris Groen},
booktitle={The Thirteenth International Conference on Learning Representations},
year={2025},
url={https://openreview.net/forum?id=CGON8Btleu}
}

@inproceedings{bao2025mindsimulator,
title={MindSimulator: Exploring Brain Concept Localization via Synthetic f{MRI}},
author={Guangyin Bao and Qi Zhang and Zixuan Gong and Zhuojia Wu and Duoqian Miao},
booktitle={The Thirteenth International Conference on Learning Representations},
year={2025},
url={https://openreview.net/forum?id=vgt2rSf6al}
}

@inproceedings{xu2019effective,
  title={Effective and Stable Neuron Model Optimization Based on Aggregated CMA-ES},
  author={Xu, Han and Shinozaki, Takahiro and Kobayashi, Ryota},
  booktitle={ICASSP 2019-2019 IEEE International Conference on Acoustics, Speech and Signal Processing (ICASSP)},
  pages={1264--1268},
  year={2019},
  organization={IEEE}
}

@inproceedings{yang2024using,
  title={Using human feedback to fine-tune diffusion models without any reward model},
  author={Yang, Kai and Tao, Jian and Lyu, Jiafei and Ge, Chunjiang and Chen, Jiaxin and Shen, Weihan and Zhu, Xiaolong and Li, Xiu},
  booktitle={Proceedings of the IEEE/CVF Conference on Computer Vision and Pattern Recognition},
  pages={8941--8951},
  year={2024}
}

@article{fan2023dpok,
  title={Dpok: Reinforcement learning for fine-tuning text-to-image diffusion models},
  author={Fan, Ying and Watkins, Olivia and Du, Yuqing and Liu, Hao and Ryu, Moonkyung and Boutilier, Craig and Abbeel, Pieter and Ghavamzadeh, Mohammad and Lee, Kangwook and Lee, Kimin},
  journal={Advances in Neural Information Processing Systems},
  volume={36},
  pages={79858--79885},
  year={2023}
}

@inproceedings{
    black2024training,
    title={Training Diffusion Models with Reinforcement Learning},
    author={Kevin Black and Michael Janner and Yilun Du and Ilya Kostrikov and Sergey Levine},
    booktitle={The Twelfth International Conference on Learning Representations},
    year={2024},
    url={https://openreview.net/forum?id=YCWjhGrJFD}
}

@inproceedings{wei2024mb2c,
  title={Mb2c: Multimodal bidirectional cycle consistency for learning robust visual neural representations},
  author={Wei, Yayun and Cao, Lei and Li, Hao and Dong, Yilin},
  booktitle={Proceedings of the 32nd ACM International Conference on Multimedia},
  pages={8992--9000},
  year={2024}
}

@inproceedings{zhang2025cognitioncapturer,
  title={CognitionCapturer: Decoding Visual Stimuli From Human EEG Signal With Multimodal Information},
  author={Zhang, Kaifan and He, Lihuo and Jiang, Xin and Lu, Wen and Wang, Di and Gao, Xinbo},
  booktitle={Proceedings of the AAAI Conference on Artificial Intelligence},
  volume={39},
  pages={14486--14493},
  year={2025},
  doi={10.1609/aaai.v39i13.33587}
}

@article{lawhern2018eegnet,
  title={EEGNet: a compact convolutional neural network for EEG-based brain--computer interfaces},
  author={Lawhern, Vernon J and Solon, Amelia J and Waytowich, Nicholas R and Gordon, Stephen M and Hung, Chou P and Lance, Brent J},
  journal={Journal of neural engineering},
  volume={15},
  number={5},
  pages={056013},
  year={2018},
  publisher={iOP Publishing}
}

@article{bashashati2016user,
  title={User-customized brain computer interfaces using Bayesian optimization},
  author={Bashashati, Hossein and Ward, Rabab K and Bashashati, Ali},
  journal={Journal of neural engineering},
  volume={13},
  number={2},
  pages={026001},
  year={2016},
  publisher={IOP Publishing}
}

@article{groen2017contributions,
  title={Contributions of low-and high-level properties to neural processing of visual scenes in the human brain},
  author={Groen, Iris IA and Silson, Edward H and Baker, Chris I},
  journal={Philosophical Transactions of the Royal Society B: Biological Sciences},
  volume={372},
  number={1714},
  pages={20160102},
  year={2017},
  publisher={The Royal Society}
}

@article{qiu2023efficient,
  title={Efficient coding of natural scenes improves neural system identification},
  author={Qiu, Yongrong and Klindt, David A and Szatko, Klaudia P and Gonschorek, Dominic and Hoefling, Larissa and Schubert, Timm and Busse, Laura and Bethge, Matthias and Euler, Thomas},
  journal={PLoS computational biology},
  volume={19},
  number={4},
  pages={e1011037},
  year={2023},
  publisher={Public Library of Science San Francisco, CA USA}
}

@inproceedings{baroni2023learning,
  title={Learning invariance manifolds of visual sensory neurons},
  author={Baroni, Luca and Bashiri, Mohammad and Willeke, Konstantin F and Antol{\'\i}k, J{\'a}n and Sinz, Fabian H},
  booktitle={NeurIPS Workshop on Symmetry and Geometry in Neural Representations},
  pages={301--326},
  year={2023},
  organization={PMLR}
}

@inproceedings{li2019controllable,
 author = {Li, Bowen and Qi, Xiaojuan and Lukasiewicz, Thomas and Torr, Philip},
 booktitle = {Advances in Neural Information Processing Systems},
 editor = {H. Wallach and H. Larochelle and A. Beygelzimer and F. d\textquotesingle Alch\'{e}-Buc and E. Fox and R. Garnett},
 publisher = {Curran Associates, Inc.},
 title = {Controllable Text-to-Image Generation},
 url = {https://proceedings.neurips.cc/paper_files/paper/2019/file/1d72310edc006dadf2190caad5802983-Paper.pdf},
 volume = {32},
 year = {2019}
}

@article{rahmani2022natural,
  title={Natural image synthesis for the retina with variational information bottleneck representation},
  author={Rahmani, Babak and Psaltis, Demetri and Moser, Christophe},
  journal={Advances in Neural Information Processing Systems},
  volume={35},
  pages={6034--6046},
  year={2022}
}

@article{lin2024sdxl_lightning,
  title={Sdxl-lightning: Progressive adversarial diffusion distillation},
  author={Lin, Shanchuan and Wang, Anran and Yang, Xiao},
  journal={arXiv preprint arXiv:2402.13929},
  year={2024}
}

@article{jang2021noise,
  title={Noise-trained deep neural networks effectively predict human vision and its neural responses to challenging images},
  author={Jang, Hojin and McCormack, Devin and Tong, Frank},
  journal={PLoS biology},
  volume={19},
  number={12},
  pages={e3001418},
  year={2021},
  publisher={Public Library of Science San Francisco, CA USA}
}

@article{alamia2023role,
  title={On the role of feedback in image recognition under noise and adversarial attacks: A predictive coding perspective},
  author={Alamia, Andrea and Mozafari, Milad and Choksi, Bhavin and VanRullen, Rufin},
  journal={Neural Networks},
  volume={157},
  pages={280--287},
  year={2023},
  publisher={Elsevier}
}

@inproceedings{
dosovitskiy2021an,
title={An Image is Worth 16x16 Words: Transformers for Image Recognition at Scale},
author={Alexey Dosovitskiy and Lucas Beyer and Alexander Kolesnikov and Dirk Weissenborn and Xiaohua Zhai and Thomas Unterthiner and Mostafa Dehghani and Matthias Minderer and Georg Heigold and Sylvain Gelly and Jakob Uszkoreit and Neil Houlsby},
booktitle={International Conference on Learning Representations},
year={2021},
url={https://openreview.net/forum?id=YicbFdNTTy}
}

@article{walker2019inception,
  title={Inception loops discover what excites neurons most using deep predictive models},
  author={Walker, Edgar Y and Sinz, Fabian H and Cobos, Erick and Muhammad, Taliah and Froudarakis, Emmanouil and Fahey, Paul G and Ecker, Alexander S and Reimer, Jacob and Pitkow, Xaq and Tolias, Andreas S},
  journal={Nature neuroscience},
  volume={22},
  number={12},
  pages={2060--2065},
  year={2019},
  publisher={Nature Publishing Group US New York}
}

@article{feather2023model,
  title={Model metamers reveal divergent invariances between biological and artificial neural networks},
  author={Feather, Jenelle and Leclerc, Guillaume and M{\k{a}}dry, Aleksander and McDermott, Josh H},
  journal={Nature Neuroscience},
  volume={26},
  number={11},
  pages={2017--2034},
  year={2023},
  publisher={Nature Publishing Group US New York}
}

@article{guggenmos2018multivariate,
  title={Multivariate pattern analysis for MEG: A comparison of dissimilarity measures},
  author={Guggenmos, Matthias and Sterzer, Philipp and Cichy, Radoslaw Martin},
  journal={Neuroimage},
  volume={173},
  pages={434--447},
  year={2018},
  publisher={Elsevier}
}

@article{gifford2022large,
  title={A large and rich EEG dataset for modeling human visual object recognition},
  author={Gifford, Alessandro T and Dwivedi, Kshitij and Roig, Gemma and Cichy, Radoslaw M},
  journal={NeuroImage},
  volume={264},
  pages={119754},
  year={2022},
  publisher={Elsevier}
}

@article{grootswagers2022human,
  title={Human EEG recordings for 1,854 concepts presented in rapid serial visual presentation streams},
  author={Grootswagers, Tijl and Zhou, Ivy and Robinson, Amanda K and Hebart, Martin N and Carlson, Thomas A},
  journal={Scientific Data},
  volume={9},
  number={1},
  pages={3},
  year={2022},
  publisher={Nature Publishing Group UK London}
}

@inproceedings{ferecatu2007interactive,
  title={Interactive search for image categories by mental matching},
  author={Ferecatu, Marin and Geman, Donald},
  booktitle={2007 IEEE 11th International Conference on Computer Vision},
  pages={1--8},
  year={2007},
  organization={IEEE}
}

@article{epstein1998cortical,
  title={A cortical representation of the local visual environment},
  author={Epstein, Russell and Kanwisher, Nancy},
  journal={Nature},
  volume={392},
  number={6676},
  pages={598--601},
  year={1998},
  publisher={Nature Publishing Group UK London}
}

@article{ye2023ip,
  title={Ip-adapter: Text compatible image prompt adapter for text-to-image diffusion models},
  author={Ye, Hu and Zhang, Jun and Liu, Sibo and Han, Xiao and Yang, Wei},
  journal={arXiv preprint arXiv:2308.06721},
  year={2023}
}

@inproceedings{li2024visual,
 author = {Li, Dongyang and Wei, Chen and Li, Shiying and Zou, Jiachen and Liu, Quanying},
 booktitle = {Advances in Neural Information Processing Systems},
 doi = {10.52202/079017-3266},
 editor = {A. Globerson and L. Mackey and D. Belgrave and A. Fan and U. Paquet and J. Tomczak and C. Zhang},
 pages = {102822--102864},
 publisher = {Curran Associates, Inc.},
 title = {Visual Decoding and Reconstruction via EEG Embeddings with Guided Diffusion},
 url = {https://proceedings.neurips.cc/paper_files/paper/2024/file/ba5f1233efa77787ff9ec015877dbd1f-Paper-Conference.pdf},
 volume = {37},
 year = {2024}
}

@article{luo2023brain,
  title={Brain diffusion for visual exploration: Cortical discovery using large scale generative models},
  author={Luo, Andrew and Henderson, Maggie and Wehbe, Leila and Tarr, Michael},
  journal={Advances in Neural Information Processing Systems},
  volume={36},
  pages={75740--75781},
  year={2023}
}

@inproceedings{
tan2025fast,
title={Fast Direct: Query-Efficient Online Black-Box Guidance for Diffusion-Model Target Generation},
author={Kim Yong Tan and Yueming Lyu and Ivor Tsang and Yew-Soon Ong},
booktitle={The Thirteenth International Conference on Learning Representations},
year={2025},
url={https://openreview.net/forum?id=OmpTdjl7RV}
}

@inproceedings{
podell2024sdxl,
title={{SDXL}: Improving Latent Diffusion Models for High-Resolution Image Synthesis},
author={Dustin Podell and Zion English and Kyle Lacey and Andreas Blattmann and Tim Dockhorn and Jonas M{\"u}ller and Joe Penna and Robin Rombach},
booktitle={The Twelfth International Conference on Learning Representations},
year={2024},
url={https://openreview.net/forum?id=di52zR8xgf}
}

@article{pierzchlewicz2023energy,
  title={Energy guided diffusion for generating neurally exciting images},
  author={Pierzchlewicz, Pawel and Willeke, Konstantin and Nix, Arne and Elumalai, Pavithra and Restivo, Kelli and Shinn, Tori and Nealley, Cate and Rodriguez, Gabrielle and Patel, Saumil and Franke, Katrin and others},
  journal={Advances in Neural Information Processing Systems},
  volume={36},
  pages={32574--32601},
  year={2023}
}

@article{minai2024miso,
  title={MiSO: Optimizing brain stimulation to create neural activity states},
  author={Minai, Yuki and Soldado-Magraner, Joana and Smith, Matthew and Yu, Byron M},
  journal={Advances in Neural Information Processing Systems},
  volume={37},
  pages={24126--24149},
  year={2024}
}

@inproceedings{kubilius2019brain,
 author = {Kubilius, Jonas and Schrimpf, Martin and Kar, Kohitij and Rajalingham, Rishi and Hong, Ha and Majaj, Najib and Issa, Elias and Bashivan, Pouya and Prescott-Roy, Jonathan and Schmidt, Kailyn and Nayebi, Aran and Bear, Daniel and Yamins, Daniel L and DiCarlo, James J},
 booktitle = {Advances in Neural Information Processing Systems},
 editor = {H. Wallach and H. Larochelle and A. Beygelzimer and F. d\textquotesingle Alch\'{e}-Buc and E. Fox and R. Garnett},
 publisher = {Curran Associates, Inc.},
 title = {Brain-Like Object Recognition with High-Performing Shallow Recurrent ANNs},
 url = {https://proceedings.neurips.cc/paper_files/paper/2019/file/7813d1590d28a7dd372ad54b5d29d033-Paper.pdf},
 volume = {32},
 year = {2019}
}

@article{seeger2004gaussian,
  title={Gaussian processes for machine learning},
  author={Seeger, Matthias},
  journal={International journal of neural systems},
  volume={14},
  number={02},
  pages={69--106},
  year={2004},
  publisher={World Scientific}
}

@inproceedings{krizhevsky2012imagenet,
 author = {Krizhevsky, Alex and Sutskever, Ilya and Hinton, Geoffrey E},
 booktitle = {Advances in Neural Information Processing Systems},
 editor = {F. Pereira and C.J. Burges and L. Bottou and K.Q. Weinberger},
 publisher = {Curran Associates, Inc.},
 title = {ImageNet Classification with Deep Convolutional Neural Networks},
 url = {https://proceedings.neurips.cc/paper_files/paper/2012/file/c399862d3b9d6b76c8436e924a68c45b-Paper.pdf},
 volume = {25},
 year = {2012}
}

@article{gu2023human,
  title={Human brain responses are modulated when exposed to optimized natural images or synthetically generated images},
  author={Gu, Zijin and Jamison, Keith and Sabuncu, Mert R and Kuceyeski, Amy},
  journal={Communications Biology},
  volume={6},
  number={1},
  pages={1076},
  year={2023},
  publisher={Nature Publishing Group UK London}
}

@inproceedings{ijcai2024p352,
  title     = {CoCoG: Controllable Visual Stimuli Generation Based on Human Concept Representations},
  author    = {Wei, Chen and Zou, Jiachen and Heinke, Dietmar and Liu, Quanying},
  booktitle = {Proceedings of the Thirty-Third International Joint Conference on
               Artificial Intelligence, {IJCAI-24}},
  publisher = {International Joint Conferences on Artificial Intelligence Organization},
  editor    = {Kate Larson},
  pages     = {3178--3186},
  year      = {2024},
  month     = {8},
  note      = {Main Track},
  doi       = {10.24963/ijcai.2024/352},
  url       = {https://doi.org/10.24963/ijcai.2024/352},
}

@article{ramesh2022hierarchical,
  title={Hierarchical Text-Conditional Image Generation with CLIP Latents},
  author={Ramesh, Aditya and Dhariwal, Prafulla and Nichol, Alex and Chu, Casey and Chen, Mark},
  journal={arXiv preprint arXiv:2204.06125},
  year={2022}
}

@inproceedings{nichol2022glide,
  title={GLIDE: Towards Photorealistic Image Generation and Editing with Text-Guided Diffusion Models},
  author={Nichol, Alexander Quinn and Dhariwal, Prafulla and Ramesh, Aditya and Shyam, Pranav and Mishkin, Pamela and Mcgrew, Bob and Sutskever, Ilya and Chen, Mark},
  booktitle={International Conference on Machine Learning},
  pages={16784--16804},
  year={2022},
  organization={PMLR}
}

@article{lang1997international,
  title={International affective picture system (IAPS): Technical manual and affective ratings},
  author={Lang, Peter J and Bradley, Margaret M and Cuthbert, Bruce N},
  journal={NIMH Center for the Study of Emotion and Attention},
  year={1997}
}

@inproceedings{machajdik2010affective,
  title={Affective image classification using features inspired by psychology and art theory},
  author={Machajdik, Jozef and Hanbury, Allan},
  booktitle={Proceedings of the 18th ACM international conference on Multimedia},
  pages={83--92},
  year={2010}
}

@article{dan2011gaped,
  title={GAPED: Geneva Affective Picture Database},
  author={Dan-Glauser, Elise S and Scherer, Klaus R},
  journal={Behavior Research Methods},
  volume={43},
  number={2},
  pages={468--477},
  year={2011},
  publisher={Springer},
  doi={10.3758/s13428-011-0064-1}
}

@inproceedings{li2022emoset,
  title={EmoSet: A Large-Scale Emotion Dataset of Diverse Situations and Emotions},
  author={Li, Mingfei and Wu, Peng and Yuan, Fan and Zhou, Jiebo and Wang, Xinlong and Tang, Jinhui},
  booktitle={Proceedings of the 30th ACM International Conference on Multimedia},
  pages={3986--3995},
  year={2022}
}

@inproceedings{he2016deep,
  title={Deep residual learning for image recognition},
  author={He, Kaiming and Zhang, Xiangyu and Ren, Shaoqing and Sun, Jian},
  booktitle={Proceedings of the IEEE conference on computer vision and pattern recognition},
  pages={770--778},
  year={2016}
}

@inproceedings{he2020momentum,
  title={Momentum contrast for unsupervised visual representation learning},
  author={He, Kaiming and Fan, Haoqi and Wu, Yuxin and Xie, Saining and Girshick, Ross},
  booktitle={Proceedings of the IEEE/CVF conference on computer vision and pattern recognition},
  pages={9729--9738},
  year={2020}
}

@inproceedings{cherti2023reproducible,
  title={Reproducible scaling laws for contrastive language-image learning},
  author={Cherti, Mehdi and Beaumont, Romain and Wightman, Ross and Wortsman, Mitchell and Ilharco, Gabriel and Gordon, Cade and Schuhmann, Christoph and Schmidt, Ludwig and Jitsev, Jenia},
  booktitle={Proceedings of the IEEE/CVF conference on computer vision and pattern recognition},
  pages={2818--2829},
  year={2023}
}

@article{
oquab2024dinov,
title={{DINO}v2: Learning Robust Visual Features without Supervision},
author={Maxime Oquab and Timoth{\'e}e Darcet and Th{\'e}o Moutakanni and Huy V. Vo and Marc Szafraniec and Vasil Khalidov and Pierre Fernandez and Daniel HAZIZA and Francisco Massa and Alaaeldin El-Nouby and Mido Assran and Nicolas Ballas and Wojciech Galuba and Russell Howes and Po-Yao Huang and Shang-Wen Li and Ishan Misra and Michael Rabbat and Vasu Sharma and Gabriel Synnaeve and Hu Xu and Herve Jegou and Julien Mairal and Patrick Labatut and Armand Joulin and Piotr Bojanowski},
journal={Transactions on Machine Learning Research},
issn={2835-8856},
year={2024},
url={https://openreview.net/forum?id=a68SUt6zFt},
note={Featured Certification}
}

@inproceedings{caron2021emerging,
  title={Emerging properties in self-supervised vision transformers},
  author={Caron, Mathilde and Touvron, Hugo and Misra, Ishan and J{\'e}gou, Herv{\'e} and Mairal, Julien and Bojanowski, Piotr and Joulin, Armand},
  booktitle={Proceedings of the IEEE/CVF international conference on computer vision},
  pages={9650--9660},
  year={2021}
}

@article{sundaram2024does,
  title={When does perceptual alignment benefit vision representations?},
  author={Sundaram, Shobhita and Fu, Stephanie and Muttenthaler, Lukas and Tamir, Netanel and Chai, Lucy and Kornblith, Simon and Darrell, Trevor and Isola, Phillip},
  journal={Advances in Neural Information Processing Systems},
  volume={37},
  pages={55314--55341},
  year={2024}
}

@article{eimer2011face,
  title={The face-sensitive N170 component of the event-related brain potential},
  author={Eimer, Martin},
  journal={The Oxford handbook of face perception},
  volume={28},
  pages={329--344},
  year={2011}
}

@article{holm2024contribution,
  title={Contribution of low-level image statistics to EEG decoding of semantic content in multivariate and univariate models with feature optimization},
  author={Holm, Eric L{\"u}tzow and Slezak, Diego Fern{\'a}ndez and Tagliazucchi, Enzo},
  journal={NeuroImage},
  volume={293},
  pages={120626},
  year={2024},
  publisher={Elsevier}
}

@inproceedings{bai2024dreamdiffusion,
  title={DreamDiffusion: High-quality EEG-to-image generation with temporal masked signal modeling and CLIP alignment},
  author={Bai, Yunpeng and Wang, Xintao and Cao, Yan-Pei and Ge, Yixiao and Yuan, Chun and Shan, Ying},
  booktitle={European Conference on Computer Vision},
  pages={472--488},
  year={2024},
  organization={Springer}
}

@inproceedings{fu2025brainvis,
  title={BrainVis: Exploring the bridge between brain and visual signals via image reconstruction},
  author={Fu, Honghao and Wang, Hao and Chin, Jing Jih and Shen, Zhiqi},
  booktitle={ICASSP 2025-2025 IEEE International Conference on Acoustics, Speech and Signal Processing (ICASSP)},
  pages={1--5},
  year={2025},
  organization={IEEE}
}

@inproceedings{de2020brain,
  title={Brain relevance feedback for interactive image generation},
  author={de la Torre-Ortiz, Carlos and Spap{\'e}, Michiel M and Kangassalo, Lauri and Ruotsalo, Tuukka},
  booktitle={Proceedings of the 33rd Annual ACM Symposium on User Interface Software and Technology},
  pages={1060--1070},
  year={2020}
}

@inproceedings{xia2024umbrae,
  title={Umbrae: Unified multimodal brain decoding},
  author={Xia, Weihao and de Charette, Raoul and Oztireli, Cengiz and Xue, Jing-Hao},
  booktitle={European Conference on Computer Vision},
  pages={242--259},
  year={2024},
  organization={Springer}
}

@article{grizou2025self,
  title={Self-Calibrating BCIs: Ranking and Recovery of Mental Targets Without Labels},
  author={Grizou, Jonathan and de la Torre-Ortiz, Carlos and Ruotsalo, Tuukka},
  journal={arXiv preprint arXiv:2506.11151},
  year={2025}
}

@inproceedings{scotti2024mindeye2,
  title={MindEye2: Shared-Subject Models Enable fMRI-To-Image With 1 Hour of Data},
  author={Scotti, Paul Steven and Tripathy, Mihir and Torrico, Cesar and Kneeland, Reese and Chen, Tong and Narang, Ashutosh and Santhirasegaran, Charan and Xu, Jonathan and Naselaris, Thomas and Norman, Kenneth A and others},
  booktitle={International Conference on Machine Learning},
  pages={44038--44059},
  year={2024},
  organization={PMLR}
}

@inproceedings{wang2024mindbridge,
  title={Mindbridge: A cross-subject brain decoding framework},
  author={Wang, Shizun and Liu, Songhua and Tan, Zhenxiong and Wang, Xinchao},
  booktitle={Proceedings of the IEEE/CVF Conference on Computer Vision and Pattern Recognition},
  pages={11333--11342},
  year={2024}
}

@inproceedings{guo2025neuro,
  title={Neuro-3D: Towards 3D visual decoding from EEG signals},
  author={Guo, Zhanqiang and Wu, Jiamin and Song, Yonghao and Bu, Jiahui and Mai, Weijian and Zheng, Qihao and Ouyang, Wanli and Song, Chunfeng},
  booktitle={Proceedings of the Computer Vision and Pattern Recognition Conference},
  pages={23870--23880},
  year={2025}
}

@article{shen2024neuro,
  title={Neuro-vision to language: Enhancing brain recording-based visual reconstruction and language interaction},
  author={Shen, Guobin and Zhao, Dongcheng and He, Xiang and Feng, Linghao and Dong, Yiting and Wang, Jihang and Zhang, Qian and Zeng, Yi},
  journal={Advances in Neural Information Processing Systems},
  volume={37},
  pages={98083--98110},
  year={2024}
}

@inproceedings{lopez2025guess,
  title={Guess What I Think: Streamlined EEG-to-Image Generation with Latent Diffusion Models},
  author={Lopez, Eleonora and Sigillo, Luigi and Colonnese, Federica and Panella, Massimo and Comminiello, Danilo},
  booktitle={ICASSP 2025-2025 IEEE International Conference on Acoustics, Speech and Signal Processing (ICASSP)},
  pages={1--5},
  year={2025},
  organization={IEEE},
  doi={10.1109/ICASSP49660.2025.10890059}
}
\bibliographystyle{iclr2026_conference}

\clearpage
\newpage
\vspace{7pt}
\section*{\Large \centering Supplementary Material: \\ 
\vspace{7pt}
{MindPilot: Closed-loop Visual Stimulation Optimization for Brain Modulation with EEG-guided Diffusion}}

\appendix

\renewcommand{\thefigure}{A.\arabic{figure}} 
\setcounter{figure}{0}

\renewcommand{\thetable}{A.\arabic{table}} 
\setcounter{table}{0}

\section{Appendix}

Appendix includes the following sections.
\begin{itemize}
    \item \textbf{Sec.~\ref{sec:implementation_details}:} \nameref{sec:implementation_details}
    \item \textbf{Sec.~\ref{sec:add_experimental_results}:} \nameref{sec:add_experimental_results}
    \item \textbf{Sec.~\ref{sec:EEG_verification}:} \nameref{sec:EEG_verification}
    \item \textbf{Sec.~\ref{sec:additional_retrieval_examples}:} \nameref{sec:additional_retrieval_examples}
    \item \textbf{Sec.~\ref{sec:additional_generation_examples}:} \nameref{sec:additional_generation_examples}
\end{itemize}

\subsection{More implementation details}
\label{sec:implementation_details}

\subsubsection{Public Datasets}
\label{sec:appendix_datasets}
We conducted our experiments using the training set of the THINGS-EEG2 dataset~\citep{gifford2022large, grootswagers2022human}, which consists of a large EEG corpus from 10 human subjects performing a visual task. The experiments used the Rapid Serial Visual Presentation (RSVP) paradigm for orthogonal target detection tasks to ensure participants' attention to the visual stimuli. All 10 participants underwent 4 equivalent experiments, resulting in 10 datasets with 16,540 unique training image conditions, each repeated 4 times, and 200 unique testing image conditions, each repeated 80 times. In total, this yielded (16,540 training image conditions \(\times\) 4 repetitions) + (200 testing image conditions \(\times\) 80 repetitions) = 82,160 image trials. The original data were recorded using a 64-channel EEG system with a 1000 Hz sampling rate. For preprocessing, the data were first downsampled to 250 Hz, and 17 channels were selected from the occipital and parietal regions, which are closely related to the visual system. These channels are O1, Oz, O2, PO7, PO3, POz, PO4, PO8, P7, P5, P3, P1, Pz, P2, P4, P6, and P8. The EEG data were then segmented into trials, spanning from 0 to 1000 ms post-stimulus onset, with baseline correction applied using the mean of the 200 ms pre-stimulus period. Multivariate noise normalization was applied to the training data~\citep{guggenmos2018multivariate}.

\subsubsection{Human experimental data acquisition}
\label{sec:add_human_exp}
\paragraph{Ethics approval and informed consent}
The human-subject protocol was reviewed and approved by the Science and Technology Ethics Committee of Southern University of Science and Technology (SUSTech; decision no.~20250292, acceptance no.~2025PES292) on June 30, 2025. The approval covers the project entitled ``Emotion-related EEG decoding, modeling, and neurofeedback regulation in naturalistic settings,'' an English translation of the project title in the approval document. Before enrollment, all participants were informed of the study purpose and procedures, including visual stimulation, non-invasive EEG recording, and the collection of subjective similarity and emotional-valence ratings. All participants provided written informed consent before data collection and were informed that participation was voluntary and that they could withdraw at any time without penalty. Participants were compensated according to the duration of their participation.

\paragraph{Participants}
Participation in the neurophysiological experiment was advertised to the university population. The mean age of the participants was 21 $\pm$ 3 years. All participants were cisgender (6 males and 4 females).

\paragraph{Mental matching task}
Mental matching is an interactive visual-search task~\citep{geman2006interactive}: participants first memorize a target image and then view candidate images while EEG is recorded. In our experiment, the server selected and displayed the target image, so the target was known to the participant and experimental apparatus. During iterative search, however, MindPilot used EEG-derived feedback rather than directly optimizing candidate images against the target image's pixels or visual embedding. The objective was to move the evoked EEG representation toward the specified target representation, not to reconstruct the target image pixel by pixel.

The mental matching task integrated real-time EEG acquisition with an image-stimulation paradigm. At the beginning of each trial, the server randomly selected a target image from the THINGS-EEG2 dataset (comprising 200 images). Participants were instructed to view and mentally encode the target image for a duration of one minute. Subsequently, they were presented with a sequence of server-generated or retrieved images, during which their EEG signals were continuously recorded. Each image was shown for 5 seconds, and a trigger signal was sent to the EEG acquisition device at image onset to accurately mark the temporal window corresponding to image perception.

\paragraph{Emotion regulation task}  

The emotion regulation task involved an online image-stimulation and self-assessment system designed to probe participants’ affective responses. Participants viewed a series of images presented by the server, either retrieved from an image pool or generated dynamically. Each image was displayed for 2 seconds, after which participants rated their emotional response using a scale ranging from 0.00 to 1.00, with higher scores indicating stronger positive affect. This setup allowed for the real-time capture of subjective emotional evaluations in response to visual stimuli. Participants were instructed as follows: “If the image evokes positive, warm, or happy feelings, please assign a higher score. If the image evokes negative, sad, or disgusting feelings, please assign a lower score. Try to quantify your feelings as precisely as possible and avoid extreme ratings unless strongly warranted.”

\paragraph{Image Set}

The emotion regulation task utilized a curated pool of 127 images. Of these, 105 were sourced from a validated Picture Database for Discrete Emotions, constructed using images drawn from established affective image sets including the International Affective Picture System (IAPS) \citep{lang1997international}, ArtPhoto \citep{machajdik2010affective}, and the Geneva Affective Picture Database (GAPED) \citep{dan2011gaped}, as well as selected online platforms. All images in this database were validated through large-scale subjective ratings. An additional 22 images were selected from EmoSet, a large-scale dataset designed for affective computing research \citep{li2022emoset}.

\paragraph{EEG recording}

The participants were seated in front of a 24'' LCD screen, with a resolution of 1920 x 1080 pixels and the refresh rate is 60 Hz. EEG data were recorded using the Neuracle NeuSen W4 wireless EEG acquisition system. A total of 59 scalp electrodes were used in accordance with the international 10–20 system, providing dense spatial coverage of frontal, central, parietal, occipital, and temporal regions. Independent reference and ground electrodes were included in the system. The EEG signals were sampled at a frequency of 250 Hz, sufficient for capturing the temporal dynamics of cognitive and affective processing.

The full set of recording sites included: Fpz, Fp1, Fp2, AF3, AF4, AF7, AF8, Fz, F1, F2, F3, F4, F5, F6, F7, F8, FCz, FC1, FC2, FC3, FC4, FC5, FC6, FT7, FT8, Cz, C1, C2, C3, C4, C5, C6, T7, T8, CP1, CP2, CP3, CP4, CP5, CP6, TP7, TP8, Pz, P3, P4, P5, P6, P7, P8, POz, PO3, PO4, PO5, PO6, PO7, PO8, Oz, O1, and O2.

\paragraph{EEG preprocessing}

Following each image stimulus, EEG signals recorded by the acquisition device were immediately transmitted to the server. The server identified event markers (triggers) within the data stream and segmented the EEG signals accordingly, aligning each segment precisely with its corresponding image presentation. Each EEG epoch, corresponding to a single image, was then subjected to real-time preprocessing to ensure low-latency signal analysis.

To accommodate the constraints of real-time processing, we did not apply computationally intensive artifact removal techniques such as Independent Component Analysis (ICA) for eliminating eye movement or muscle artifacts. Instead, a lightweight preprocessing pipeline was implemented using Python’s SciPy library, including a notch filter centered at 50 Hz to remove power line interference, a 1–100 Hz band-pass filter to retain relevant neural activity, and optional signal resampling. Additionally, baseline correction was applied using the final 250 ms of the pre-stimulus period to minimize low-frequency drifts and slow fluctuations.

\begin{algorithm}[t]
\caption{Closed-loop Interactive Search Algorithm}
\label{alg:interactive_search}
\begin{algorithmic}[1]
\State \textbf{Initialize:} Database $\Omega = \{u_1, ..., u_N\}$. Scores $S_0(u_j)$ for all $j \in \{1, ..., N\}$.
\State Set uniform selection probability as the initial scores $S_0(u_j) = P_0(u_j) = \frac{1}{N}$. Set hyperparameters $\alpha, \beta$.
\State Set the number of best items to consider, $k \ge 1$.
\Repeat
    \State \textbf{Action Selection:} Sample a subset of images $U_t \subset \Omega$ based on probabilities $P_t$.
    \State \textbf{Reward Calculation:} For each image $u_i \in U_t$, calculate its similarity score:
    \State $r_i \leftarrow \text{sim}(f(g(u_i)), y_{\text{target}})$
    \State \textbf{Find Top-k Best:} Identify the set of indices $I_{\text{best}}$ for the top $k$ images in $U_t$ with the highest scores.
    
    \State \textbf{Step A: Direct Reward Update}
    \State Compute intermediate scores $S'_t$ for all $u_i \in U_t$:
    \State $S'_{t}(u_i) \leftarrow 
    \begin{cases}
        (1-\alpha) \cdot S_{t}(u_i) + \alpha \cdot r_i & \text{if } i \in I_{\text{best}} \\
        S_{t}(u_i) & \text{if } i \notin I_{\text{best}}
    \end{cases}$
    
    \State \textbf{Step B: Spreading Update (Aggregated from Top-k)}
    \State Compute updated scores $S_{t+1}$ for all $u_j \in \Omega$ by averaging the spread from all top-k images:
    \State $S_{t+1}(u_j) \leftarrow (1-\beta) \cdot S'_{t}(u_j) + {\beta} \cdot \frac{1}{k} \sum_{i \in \mathcal{I}_{\text{best}}} \left( \bm{S}'_{t}(\bm{u}_i) \cdot \frac{\exp(s(\bm{u}_i, \bm{u}_j))}{\sum_{l=1}^{N}\exp(s(\bm{u}_i, \bm{u}_l))} \right)$
    \Comment{$s$ is CLIP embedding similarity.}
    
    \State \textbf{Update Action Probabilities}
    \State Update the probability $P_{t+1}$ using softmax for the next iteration:
    \State $P_{t+1}(u_j) \leftarrow \frac{\exp(S_{t+1}(u_j))}{\sum_{l=1}^{N}\exp(S_{t+1}(u_l))}$
    
    \State $t \leftarrow t+1$
\Until{convergence criteria met}
\State \textbf{Return:} $\arg\max_{u_j \in \Omega} S_t(u_j)$
\end{algorithmic}
\end{algorithm}

\begin{algorithm}[t]
\caption{Closed-loop Heuristic Generation Algorithm}
\label{alg:heuristic_generation}
\begin{algorithmic}[1]
\State \textbf{Initialize:} 
Database $\Omega_0$, scores $S_0(u_i)$ for all $i \in \{1, ..., N\}$, probabilities $P_0(u_i) = S_0(u_i) = \frac{1}{N}$.
\State Parameters: parent population size $N_p$, top-$k$ offspring count $k$, hyperparameters $\alpha, \beta$.
\State Operators: parent selector $\text{SelectParents}(\cdot)$, embedding crossover $\text{Crossover}(\cdot, \cdot)$, image generator $\text{Generate}(\cdot)$.

\Repeat
    \State $U_{\text{parents}} \leftarrow \text{Sample}(\Omega_t, P_t, N_p)$ \Comment{Select parents based on global probabilities}
    
    \State $\begin{aligned}
        U_{\text{offspring}} \leftarrow \{\,&\text{Generate}(\text{Crossover}(F(u_a),F(u_b)))\\
        &\mid (u_a,u_b)\sim\text{SelectParents}(U_{\text{parents}},S_t)\}_{i=1}^{N_p}
    \end{aligned}$
    
    \State Calculate rewards $\{r_j \mid u_j \in U_{\text{offspring}}\}$ where $r_j \leftarrow sim(f(g(u_j)), y_{\text{target}})$.
    
    \State $\Omega_{t+1} \leftarrow \Omega_t \cup U_{\text{offspring}}$ \Comment{Integrate new offspring into the database}
    \State Extend $S_t$ to cover $\Omega_{t+1}$ with initial scores for new images.

    \State $\mathcal{I}_{\text{best}} \leftarrow \arg\text{topk}_{j} \{r_j \mid u_j \in U_{\text{offspring}}\}$ \Comment{Identify best new offspring}
    
    \State Compute intermediate scores $S'_{t}$ for all $u_i \in \Omega_{t+1}$:
    \State $S'_{t}(u_i) \leftarrow 
    \begin{cases} 
        (1-\alpha) \cdot S_{t}(u_i) + \alpha \cdot r_i & \text{if } i \in \mathcal{I}_{\text{best}} \\
        S_{t}(u_i) & \text{if } i \notin \mathcal{I}_{\text{best}}
    \end{cases}$

    \State Compute final scores $S_{t+1}$ for all $u_j \in \Omega_{t+1}$:
    \State $S_{t+1}(u_j) \leftarrow (1-\beta)S'_{t}(u_j) + \frac{\beta}{k} \sum_{j \in \mathcal{I}_{\text{best}}} \left( S'_{t}(u_i) \cdot \frac{\exp(s(u_i, u_j))}{\sum_{l \in \Omega_{t+1}}\exp(s(u_i, u_l))} \right)$

    \State Update global probabilities $P_{t+1}$ for all $u_j \in \Omega_{t+1}$:
    \State $P_{t+1}(u_j) \leftarrow \frac{\exp(S_{t+1}(u_j))}{\sum_{l \in \Omega_{t+1}}\exp(S_{t+1}(u_l))}$
    
    \State $t \leftarrow t+1$
\Until{convergence criteria met}
\State \textbf{Return:} $\arg\max_{u_j \in \Omega_t} S_t(u_j)$
\end{algorithmic}
\end{algorithm}

\subsubsection{Interactive Searching Algorithm Pipeline}
\label{sec:Retrieval_pipeline}

We provide a more detailed description of algorithm~\ref{alg:interactive_search}. The algorithm begins by initializing equal selection probabilities for each image in the candidate set, denoted as $P_{0}(u_j)=\frac{1}{N}$, where $N$ is the total number of images in the retrieval set. This initialization with equal probabilities reflects the absence of prior information, serving as an exploratory phase. In each iteration, a subset of images $U_t = \{ u_1, u_2, \dots, u_N \} $ is selected from the candidate images space $\Omega$  based on the current selection probabilities $P_t(u)$. 

\begin{figure}[ht]
    \centering
    \includegraphics[width=0.6\linewidth]{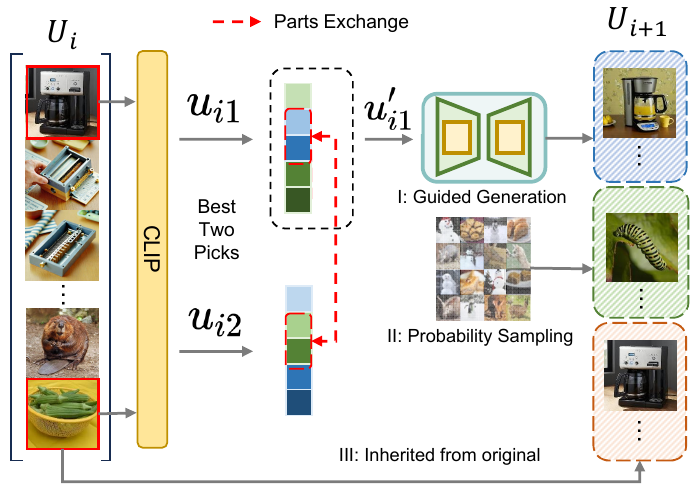}
    \caption{Generating subsequent images based on the current round is achieved through crossover, mutation, and a guided diffusion model. Both crossover and mutation operations preserve the relative ordering of CLIP features, thereby maintaining their semantic coherence.}    
    \label{fig:figGA}
\end{figure}

For each image $u^k_+$ in the subset $U_{t}$ the algorithm computes a similarity score $sim\langle u^k_+, u_{\text{target}} \rangle$ by comparing the image’s representation with the target. This similarity score acts as an immediate \textbf{reward} within the framework. The maximum similarity score among the subset is identified as a measure of the effectiveness of the current action. If $sim_{\max} $ does not meet a predefined $threshold$, the reward is considered insufficient, and the algorithm returns to the image selection step, effectively trying a new action within the same state.  If $sim_{\max} $ meets or exceeds the $threshold$, the algorithm proceeds to identify the images with the highest similarity scores. 

As for each image $u^k_+$ in $U_t$, its selection probability $P_{t+1}(u^k_+)$ is updated by multiplying with a constant factor, representing a policy improvement step that prioritizes images likely to yield higher rewards. After updating, a Softmax function is applied to normalize the probabilities, focusing selection weight on images more similar to the target. This normalization step reflects the transition to a new state with an updated policy. The iteration continues, with the algorithm transitioning through states by selecting new subsets based on the refined probabilities, until $sim_{\max}$ reaches $threshold$ or reach the upper limit of the number of iteration rounds. At this point, the loop terminates, as the algorithm has successfully identified an optimized subset of images that maximizes the similarity reward to the target.

\subsubsection{Heuristic Generation Algorithm pipeline}
\label{sec:Generation_pipeline}
We provide a more detailed description of algorithm~\ref{alg:heuristic_generation}. As illustrated in Fig.~\ref{fig:figGA}, each image set consists of three parts:

\begin{itemize}
    \item \textbf{Latent Space Crossover and Generation:} This phase integrates evolutionary operators with the diffusion generation process. Two parent instances, denoted as \( u_a \) and \( u_b \), are selected via roulette wheel sampling based on their fitness. We apply a randomized crossover operation to their latent embeddings, employing a stochastic starting index to maximize variability. The resulting hybrid embedding serves as the conditioning input for the diffusion model. This approach effectively expands the search space (exploration) while preserving the high-fidelity semantic structures of the parent nodes (exploitation).
    \item \textbf{Novelty Injection:} To mitigate mode collapse and maintain population diversity, we introduce an exploratory sampling step. New candidates are drawn from the original dataset, strictly excluding instances utilized in prior iterations. This ensures the continuous introduction of novel semantic elements, preventing the optimization process from stalling in local optima.
    \item \textbf{Elitist Preservation:} We employ an elitist strategy by directly propagating the selected parent instances \( u_a \) and \( u_b \) to the subsequent generation. This preservation mechanism ensures monotonic improvement of the population quality and stabilizes the evolutionary trajectory.
\end{itemize}
By combining these three parts, we obtain a new image set for the next iteration.

\subsection{Additional Experimental Results}
\label{sec:add_experimental_results}

\subsubsection{ATM-S as an Oracle Reference}
\label{sec:appendix_upper_bound}

For context, we report the direct-reconstruction performance of ATM-S as an oracle reference. The two methods use different protocols:

\textbf{MindPilot (Our Method):} Operates via iterative, closed-loop feedback. It searches for an image that matches a target brain state without ever accessing the ground-truth EEG signal corresponding to the target image. 

\textbf{ATM-S oracle reference:} ATM-S uses the ground-truth EEG signal corresponding to a known target image to reconstruct visual content directly, bypassing MindPilot's iterative search setting. Its result is therefore a reference ceiling under a different protocol, not a strict theoretical upper bound on MindPilot.

We selected ATM-S because it achieved the strongest results among the EEG-to-image encoders evaluated on the THINGS-EEG2 retrieval task. As shown in Table~\ref{tab:encoder_benchmark}, ATM-S has the highest reported value in each listed metric for this comparison, making it a useful reference ceiling while preserving the distinction between direct reconstruction and iterative optimization.

\begin{table*}[h]
\centering
\caption{\textbf{EEG-to-image encoder performance on THINGS-EEG2 (in-subject).} Among the listed decoding baselines, ATM-S has the highest reported retrieval accuracy across the displayed $k$-way metrics and is therefore used as an oracle/reference ceiling.}
\label{tab:encoder_benchmark}
\resizebox{\textwidth}{!}{%
\begin{tabular}{l|ccccc}
\toprule
\textbf{Model} & \textbf{2-way Top-1} & \textbf{4-way Top-1} & \textbf{10-way Top-1} & \textbf{200-way Top-1} & \textbf{200-way Top-5} \\
\midrule
NICE\citep{song2024decoding} & 93.23 & 83.93 & 69.22 & 21.67 & 51.34 \\
EEGNetV4\citep{lawhern2018eegnet} & 91.42 & 80.21 & 63.37 & 16.84 & 42.58 \\
CogCap\citep{zhang2025cognitioncapturer} & 93.15 & 82.85 & 69.35 & 22.05 & 51.60 \\
MB2C\citep{wei2024mb2c} & 78.40 & 62.25 & 43.75 & 8.85 & 25.20 \\
MindEyeV2\citep{scotti2024mindeye2} & 92.50 & 82.80 & 66.10 & 23.80 & 50.25 \\
\midrule
\textbf{ATM-S\citep{li2024visual}} & \textbf{94.70} & \textbf{86.73} & \textbf{74.00} & \textbf{26.85} & \textbf{57.21} \\
\bottomrule
\end{tabular}%
}
\end{table*}

\subsubsection{Hyperparameter Sensitivity Analysis}
\label{sec:appendix_hyperparams}

In the Interactive Searching experiments, we utilized a fixed setting for the hyperparameters $\alpha$ (moving factor) and $\beta$ (reward propagation factor) based on initial pilot observations. To evaluate sensitivity to these choices, we performed a grid search on the validation set over $[0.1, 0.9]$ for both $\alpha$ and $\beta$. The results are summarized in Table~\ref{tab:ablation_alpha_beta}.

The best validation-set performance in this grid is achieved when $\alpha=0.1$ and $\beta=0.1$, yielding a similarity score of $0.6586 \pm 0.0879$. Lower values of $\beta$ (controlling reward propagation spread) tend to yield higher scores in this experiment, suggesting that more localized reward feedback may be beneficial for this task. The grid therefore documents sensitivity to the selected hyperparameters rather than a universal performance bound.

\begin{table*}[h]
\centering
\caption{\textbf{Ablation study on hyperparameters $\alpha$ and $\beta$.} The performance is evaluated on the validation set. The values represent the target score (mean $\pm$ std). The best performance is achieved at $\alpha=0.1, \beta=0.1$, indicating that finer-grained updates and localized reward propagation favor optimization.}
\label{tab:ablation_alpha_beta}
\resizebox{\textwidth}{!}{%
\begin{tabular}{c|ccccc}
\toprule
\textbf{\diagbox{$\alpha$}{$\beta$}} & \textbf{0.1} & \textbf{0.3} & \textbf{0.5} & \textbf{0.7} & \textbf{0.9} \\
\midrule
\textbf{0.1} & $\mathbf{0.6586 \pm 0.0879}$ & $0.6336 \pm 0.0584$ & $0.6284 \pm 0.0644$ & $0.6167 \pm 0.0658$ & $0.6224 \pm 0.0600$ \\
\textbf{0.3} & $0.6586 \pm 0.0879$ & $0.6336 \pm 0.0584$ & $0.6284 \pm 0.0644$ & $0.6203 \pm 0.0709$ & $0.6224 \pm 0.0600$ \\
\textbf{0.5} & $0.6586 \pm 0.0879$ & $0.6336 \pm 0.0584$ & $0.6264 \pm 0.0667$ & $0.6203 \pm 0.0709$ & $0.6224 \pm 0.0600$ \\
\textbf{0.7} & $0.6586 \pm 0.0879$ & $0.6292 \pm 0.0592$ & $0.6264 \pm 0.0667$ & $0.6203 \pm 0.0709$ & $0.6224 \pm 0.0600$ \\
\textbf{0.9} & $0.6586 \pm 0.0879$ & $0.6298 \pm 0.0599$ & $0.6264 \pm 0.0667$ & $0.6217 \pm 0.0700$ & $0.6224 \pm 0.0600$ \\
\bottomrule
\end{tabular}%
}
\end{table*}

In the Tab.\ref{tab:ablation_alpha_beta}, we found that the $\beta$ parameter has a significant impact on the results. This might be because when we implemented the codes, we included a standardized penalty term for direct rewards, in order to ensure a stable acquisition of rewards, which made finding the distribution of potential target rewards more important than a single reward.

\subsubsection{Necessity of EEG-Driven Optimization}
\label{sec:appendix_clip_baseline}

To assess the contribution of EEG-derived feedback, we compared MindPilot with a random-guidance null condition and a target-image oracle reference. In this control experiment, the optimizer does not receive the target image and instead uses the target brain feature as its goal. We report three setups:

\begin{enumerate}
    \item \textbf{Random Image Guidance (Lower Bound/Null):} No optimization is performed. Removing the EEG renders the system without any optimization direction. This establishes the baseline performance.
    \item \textbf{EEG Feature Guidance (MindPilot):} Optimization is driven solely by EEG feedback, where the target is the specific brain state.
    \item \textbf{Target Image Guidance (Oracle reference):} Optimization is driven directly by the ground-truth target image's CLIP features. This condition uses information unavailable to MindPilot and is reported as a protocol-specific oracle reference.
\end{enumerate}

The quantitative results are presented in Tab.~\ref{tab:clip_baseline}. MindPilot outperforms the random baseline across the reported metrics. For instance, in the closed-loop setting, MindPilot achieves a CLIP score of $0.7065 \pm 0.0537$ compared with $0.5746 \pm 0.0206$ for random guidance. This comparison supports the contribution of EEG-derived feedback in the evaluated setup; target-image guidance remains an oracle reference under a different information condition.

\begin{table*}[h]
\centering
\caption{\textbf{Control experiments with different guiding targets.} Comparison of MindPilot (EEG guidance), target-image guidance (oracle reference), and random guidance (null condition).}
\label{tab:clip_baseline}
\resizebox{\textwidth}{!}{%
\begin{tabular}{lcccccc}
\toprule
\multirow{2}{*}{\textbf{Type}} & \multicolumn{2}{c}{\textbf{EEG feature guidance}} & \multicolumn{2}{c}{\textbf{Target image guidance}} & \multicolumn{2}{c}{\textbf{Random image guidance}} \\
 & \multicolumn{2}{c}{\textbf{(MindPilot)}} & \multicolumn{2}{c}{\textbf{(Ceiling)}} & \multicolumn{2}{c}{\textbf{(Null)}} \\

\cmidrule(lr){2-3} \cmidrule(lr){4-5} \cmidrule(lr){6-7}
 & EEG score & CLIP score & EEG score & CLIP score & EEG score & CLIP score \\
\midrule

Offline (200 pairs) & 
0.5369 $\pm$ 0.0160 & 0.6580 $\pm$ 0.0492 & 
\textbf{0.5452 $\pm$ 0.0122} & \textbf{0.8464 $\pm$ 0.0553} & %
0.5223 $\pm$ 0.0160 & \textbf{0.5849 $\pm$ 0.0286} \\
\midrule
Closed-loop (10 loops) & 
\textbf{0.5461 $\pm$ 0.0141} & \textbf{0.7065 $\pm$ 0.0537} & 
0.5440 $\pm$ 0.0121 & 0.7838 $\pm$ 0.0699 & %
\textbf{0.5236 $\pm$ 0.0132} & 0.5746 $\pm$ 0.0206 \\
\bottomrule
\end{tabular}%
}
\end{table*}

\subsubsection{Comparison with Black-box Optimizers and Efficiency Analysis}
\label{sec:appendix_optimizer_comparison}

In this section, we provide a detailed theoretical justification for choosing a GP-based ``Surrogate Gradient'' approach (MindPilot) over standard Bayesian Optimization (BO) and present empirical comparisons regarding convergence speed and query efficiency.

\paragraph{Pseudo Model vs. BO} We explicitly chose a GP-based ``Surrogate Gradient'' approach rather than standard BO for two primary theoretical reasons, particularly given the high-dimensional nature of the generative latent space:

\textbf{The Curse of Dimensionality:} Standard BO relies on optimizing an acquisition function (e.g., Upper Confidence Bound (UCB) or Expected Improvement (EI)) over the input space. This process becomes computationally intractable and sample-inefficient in the high-dimensional latent space of diffusion models (e.g., $64 \times 64 \times 4 \approx 16,384$ dimensions). Even when optimizing within the same CLIP embedding space (with $1024$ dimensions) of Mindpilot, convergence is still extremely difficult. MindPilot circumvents this by using the proxy to guide gradient estimation rather than performing global optimization over the raw input space.

\textbf{Local vs. Global Search:} Standard BO aims for global optimization, which requires extensive exploration and is expensive in terms of query budget. In contrast, MindPilot uses GP to model the \textit{local} landscape around the current generation to estimate an update direction. This design combines gradient-style guidance with GP-based modeling of EEG uncertainty and noise.

To examine the performance--speed trade-off, we compared MindPilot (Pseudo-model) with representative reinforcement-learning methods (DDPO, DPOK, D3PO) and black-box optimizers (standard BO and CMA-ES). We evaluated all methods using query budgets of 5, 10, 50, and 200 samples. For the assessment stage, we selected the minimum of 5 or (num\_budget // 2) images for the evaluation of EEG scores. Running time represents the wall-clock time required to complete the optimization process. Table~\ref{tab:optimizer_efficiency} reports the resulting trade-offs.
\begin{enumerate}
    \item \textbf{vs. RL Baselines (DDPO/DPOK/D3PO):} MindPilot converges orders of magnitude faster. For example, at 200 samples, MindPilot (Offline) takes $\approx 73$s, whereas DDPO takes $\approx 1279$s and D3PO takes $\approx 1614$s.
    \item \textbf{vs. Black-box Optimizers (BO/CMA-ES):} While BO is fast at low sample counts, its computational cost scales poorly with history size due to matrix inversion (cubic complexity). At 200 samples, MindPilot (Offline) is not only faster ($\approx 73$s vs. $229$s for BO) but also achieves a higher EEG score ($0.5384$ vs. $0.5242$), validating its effectiveness in high-dimensional optimization.
\end{enumerate}

\begin{table*}[ht]
\centering
\caption{\textbf{Efficiency and Performance Benchmark.} Comparison of EEG Scores (Higher is better) and Running Time (Lower is better) across different optimization methods. MindPilot demonstrates trade-off between efficiency and scalability, especially at larger sample sizes.}
\label{tab:optimizer_efficiency}
\resizebox{\textwidth}{!}{%
\begin{tabular}{lcccccccc}
\toprule
\multirow{2}{*}{\textbf{Method}} & \multicolumn{2}{c}{\textbf{5 Samples}} & \multicolumn{2}{c}{\textbf{10 Samples}} & \multicolumn{2}{c}{\textbf{50 Samples}} & \multicolumn{2}{c}{\textbf{200 Samples}} \\
\cmidrule(lr){2-3} \cmidrule(lr){4-5} \cmidrule(lr){6-7} \cmidrule(lr){8-9}
 & EEG Score $\uparrow$ & Time (s) $\downarrow$ & EEG Score $\uparrow$ & Time (s) $\downarrow$ & EEG Score $\uparrow$ & Time (s) $\downarrow$ & EEG Score $\uparrow$ & Time (s) $\downarrow$ \\
\midrule
DDPO\citep{black2024training} & 0.5125 $\pm$ 0.0139 & 107.82 $\pm$ 19.42 & 0.5095 $\pm$ 0.0097 & 220.27 $\pm$ 17.29 & 0.5154 $\pm$ 0.0126 & 683.42 $\pm$ 38.87 & 0.5125 $\pm$ 0.0136 & 1279.59 $\pm$ 87.56 \\
DPOK\citep{fan2023dpok} & 0.5093 $\pm$ 0.0121 & 116.61 $\pm$ 21.75 & 0.5138 $\pm$ 0.0124 & 221.89 $\pm$ 12.14 & 0.5108 $\pm$ 0.0154 & 692.70 $\pm$ 40.37 & 0.5101 $\pm$ 0.0174 & 1332.15 $\pm$ 164.56 \\
D3PO\citep{yang2024using} & 0.5138 $\pm$ 0.0162 & 117.70 $\pm$ 9.93 & 0.5113 $\pm$ 0.0154 & 285.38 $\pm$ 6.56 & \textbf{0.5500 $\pm$ 0.0156} & 486.90 $\pm$ 109.96 & 0.5192 $\pm$ 0.0104 & 1614.16 $\pm$ 128.63 \\
\midrule
BO\citep{bashashati2016user} & \textbf{0.5228 $\pm$ 0.0096} & \textbf{5.35 $\pm$ 0.34} & \underline{0.5247 $\pm$ 0.0058} & \underline{11.11 $\pm$ 0.45} & 0.5222 $\pm$ 0.0087 & 53.05 $\pm$ 3.57 & 0.5242 $\pm$ 0.0111 & 229.06 $\pm$ 9.57 \\
CMA-ES\citep{xu2019effective} & \underline{0.5224 $\pm$ 0.0076} & \underline{5.40 $\pm$ 0.42} & 0.5209 $\pm$ 0.0086 & \textbf{9.26 $\pm$ 0.54} & 0.5195 $\pm$ 0.0087 & \underline{49.76 $\pm$ 2.92} & 0.5227 $\pm$ 0.0096 & 239.26 $\pm$ 14.03 \\
\midrule
MindPilot (Offline) & 0.5222 $\pm$ 0.0067 & 17.94 $\pm$ 1.61 & \textbf{0.5264 $\pm$ 0.0077} & 20.56 $\pm$ 2.87 & 0.5291 $\pm$ 0.0058 & \textbf{27.75 $\pm$ 4.89} & \textbf{0.5384 $\pm$ 0.0114} & \textbf{73.62 $\pm$ 19.02} \\
MindPilot (Closed-loop) & 0.5208 $\pm$ 0.0107 & 39.63 $\pm$ 7.37 & 0.5208 $\pm$ 0.0107 & 42.87 $\pm$ 9.71 & \underline{0.5327 $\pm$ 0.0091} & 64.67 $\pm$ 10.92 & \underline{0.5341 $\pm$ 0.0071} & \underline{219.78 $\pm$ 40.33} \\
\bottomrule
\end{tabular}%
}
\end{table*}

\subsubsection{Additional reconstructed images results}
\label{sec:reconst_figs}
\begin{figure}[htbp]
    \centering
    \includegraphics[width=1.0\linewidth]{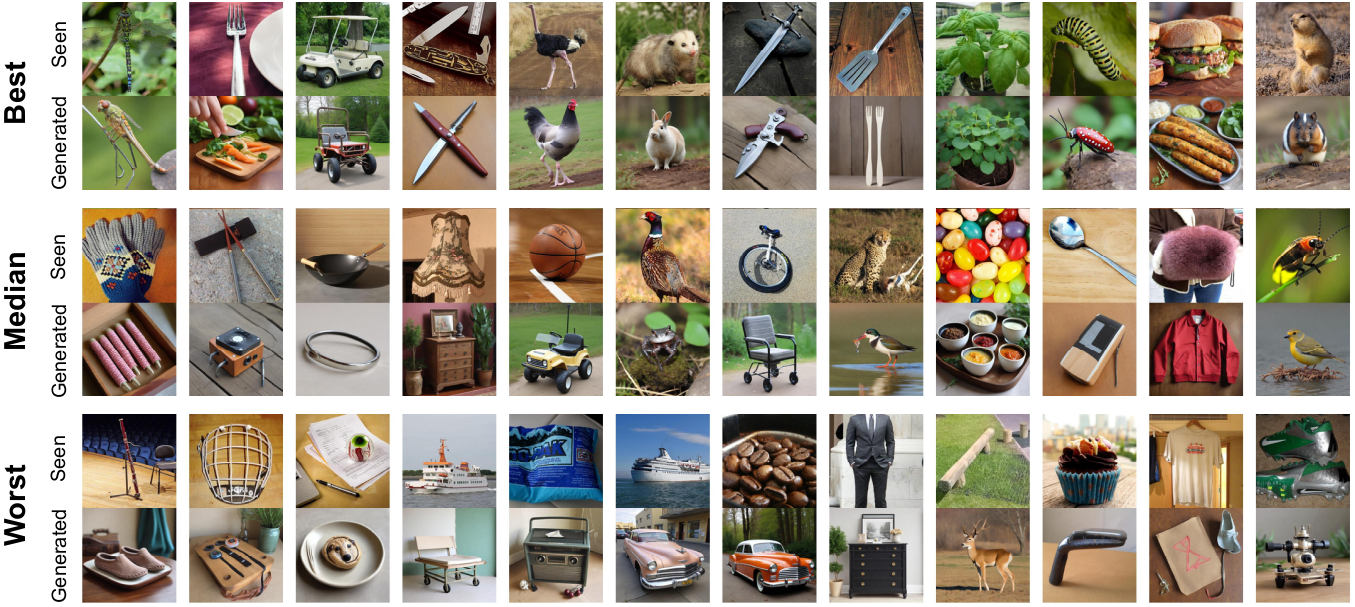}
    \caption{\textbf{Examples of EEG semantic feature-guided visual reconstruction}. From top to bottom, we exhibit the best, median, and worst 12 generated images from Subject-01, respectively. We show the images subjects had seen and the generated images.}
    \label{fig:reconst_figs}
\end{figure}

Fig.~\ref{fig:reconst_figs} visualizes the best, medium, and worst examples of our image generation results. To obtain these results, we used EEG data from Subject-01 viewing 200 test images. Each image in the test set was sequentially used as a target for an iterative optimization process. We then evaluated the generation quality by calculating the cosine similarity of CLIP embeddings between the generated and original images, selecting the 12 best, 12 medium, and 12 worst-performing examples. 

In the best group, the generated images are highly consistent with the original images in both high-level semantics and low-level visual features. In the medium group, the generated images successfully capture the core semantics, but some low-level visual details are distorted or lost. In the worst group, the generated images fail to preserve either the semantic or the low-level features of the original images.

\subsubsection{Human Experimental Correlation Analysis}
We recruited 10 participants (6 males, 4 females). The group-level trajectory analyses in Fig.~\ref{fig:human_exps}D and Fig.~\ref{fig:human_exp_config_results_emotion}C use all 10 participants. Four participants (3 males, 1 female) whose recordings were classified as having poor data quality were excluded from the correlation analyses in Fig.~\ref{fig:human_exps}E--F, leaving 6 participants (3 males, 3 females). The available study materials do not document a quantitative exclusion threshold; this limitation is stated to avoid implying that the six-participant subset was used for all analyses.

\begin{figure}[htbp]
    \centering
    \includegraphics[width=1.0\linewidth]{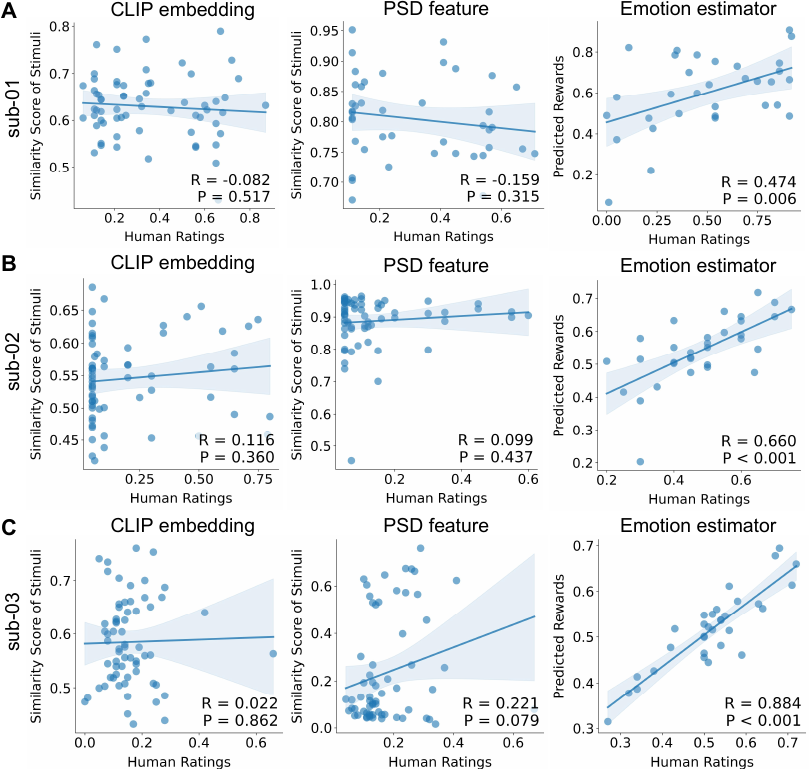}
    \caption{\textbf{Correlations analysis for S1-S3.} (Left and Middle) Correlations between the similarity score of stimulus and human rating in the mental matching task in Target 1. (Right) Correlations between the predicted rewards from the estimator and human ratings in the emotion regulation task in Target 1.}   
\end{figure}

\begin{figure}[htbp]
    \centering
    \includegraphics[width=1.0\linewidth]{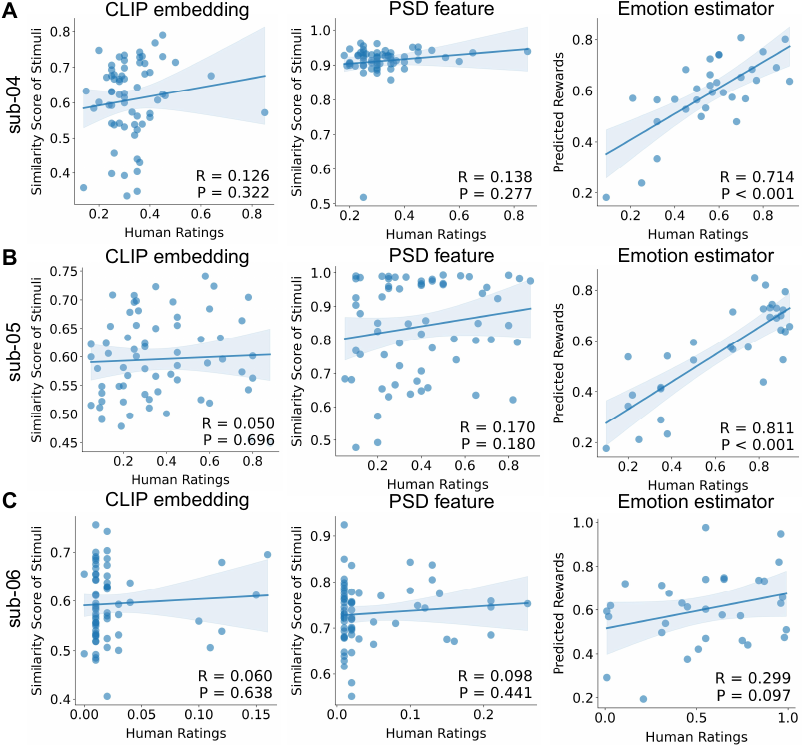}
    \caption{\textbf{Correlations analysis for S4-S6.} (Left and Middle) Correlations between the similarity score of stimulus and human rating in the mental matching task in Target 1. (Right) Correlations between the predicted rewards from the estimator and human ratings in the emotion regulation task in Target 1. }  
    \label{fig:human_all_scatters_4_6}
\end{figure}

\subsubsection{Iteration Improvement From Different Subjects}

\begin{figure}[htbp]
    \centering
    \includegraphics[width=1.0\linewidth]{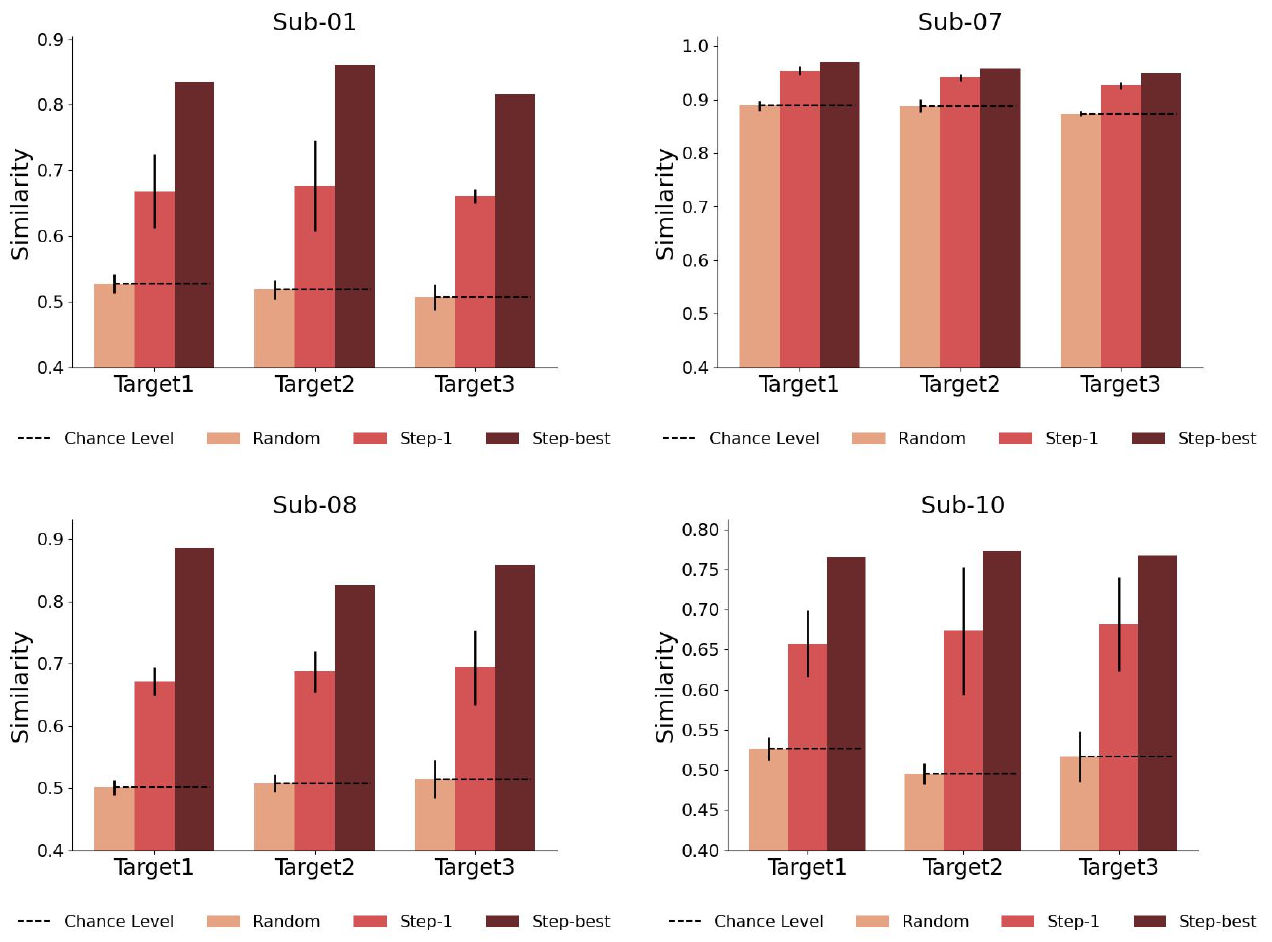}
    \caption{\textbf{Comparison of improved performance by different targets.} We present the similarity scores of EEG features generated by random stimulation, open-loop stimulation (step 1), and step-best stimulation, in comparison to the target features. Each subject randomly selected 3 images from the searching space as target images.}    
    \label{fig:appendix_targets_bar_sub}
\end{figure}

Based on the conclusions drawn from Fig.~\ref{fig:appendix_retrieval_bar}, we employ the pre-trained AlexNet end-to-end model as the black-box proxy model and use ATM-S, which is based on semantic similarity score (both the training and testing signals are synthesized), to obtain semantic representations aligned with 1×1024 CLIP image features. The experimental design involves randomly selecting 50 categories, resulting in a searching space of 50 × 12 = 600 images. Specifically, we present the iterative performance improvements for three different targets randomly selected from the test set, with results reported for Subjects 1, 7, 8, and 10. As shown in Fig.~\ref{fig:appendix_targets_bar_sub}, we calculate the EEG feature similarity of Subject 1, 7, 8, and 10 at random, step-1, and step-best in the iterative process respectively.

\subsubsection{Performance of Different Target Images Across Subjects}

\begin{figure}[h]
    \centering
    \includegraphics[width=1.0\linewidth]{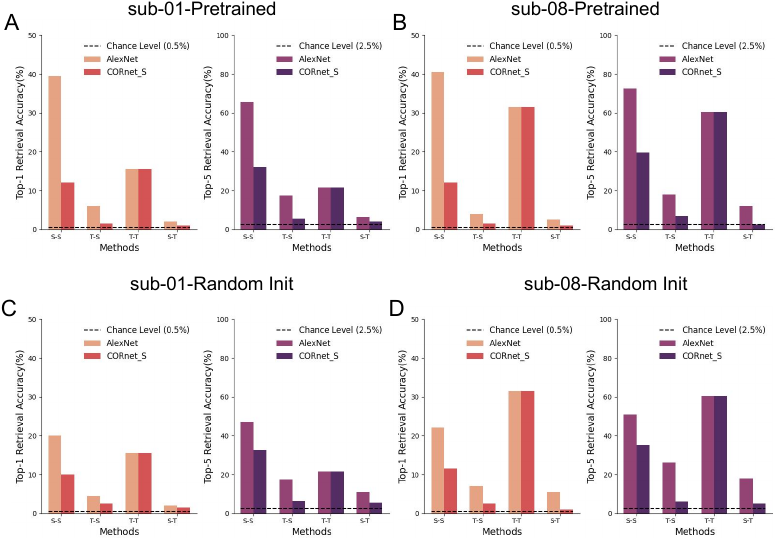}
    \caption{Retrieval accuracy under different training and test datasets. Zero-shot retrieval performance of EEG data from different sources in Subject 1 and Subject 8 using ATM-S in different Settings. AlexNet and CORnet-S used in the first row were both pre-trained end-to-end models, and the second row was randomly initialized end-to-end.}    
    \label{fig:appendix_retrieval_bar}
\end{figure}

We report the results of iterative optimization using different targets in two different cases. The results for each subject are shown, along with the average percentage improvement across 5 random seeds. For the EEG semantic feature case, we determined that training and testing with synthetic EEG yielded the highest accuracy based on the retrieval performance shown in Fig.~\ref{fig:appendix_retrieval_bar}. For the PSD feature, we selected 3 images using the method described in Section~\ref{sec:experiments} and supplemented the iterative improvement performance.

\subsection{Validity Verification of Synthetic EEG}
\label{sec:EEG_verification}
To evaluate the performance of our EEG encoding models, we compare the synthetic EEG signals generated by two deep neural networks (DNNs)—AlexNet and CORnet-S—with real EEG data. Here's a step-by-step breakdown of how we processed and compared the data.

We selected 17 specific channels from the original 63-channel EEG dataset, focusing on those most relevant to visual processing. It ensured that we focused on neural regions most directly involved in responding to the visual stimuli. For each stimulus, we averaged the EEG signals across all trials, resulting in a representative dataset for each stimulus. This reduced the dimensionality of the data, making it easier to compare with synthetic data. We used a pretrained end-to-end encoding model to generate synthetic EEG signals based on the visual stimuli. The model captures the mapping between the visual input and the resulting EEG signals using deep neural networks. These synthetic signals represent the neural responses predicted by the model in response to the stimuli.

\begin{table}[h]
    \centering
    \small 
    \setlength{\tabcolsep}{3pt} 
    \renewcommand{\arraystretch}{1.0} 
    \caption{MSE Values for synthesized EEG}
    \label{tab:mse_compact}
    \begin{tabular}{cccccc}
        \toprule
        \multirow{2}{*}{\textbf{Subject}} & \multicolumn{2}{c}{\textbf{Pretrained}} & \multicolumn{2}{c}{\textbf{Random Init}} & \multirow{2}{*}{\textbf{Average}} \\ 
        \cmidrule(lr){2-3} \cmidrule(lr){4-5}
        & \textbf{AlexNet} & \textbf{CORnet-S} & \textbf{AlexNet} & \textbf{CORnet-S} & \\ 
        \midrule
        Sub-01 & 0.1095 & 0.1126 & 0.1161 & 0.0994 & 0.1094 \\
        Sub-02 & 0.0764 & 0.0788 & 0.0840 & 0.0994 & 0.0847  \\
        Sub-03 & 0.0787 & 0.0806 & 0.0816 & 0.0910 & 0.0830 \\
        Sub-04 & 0.0652 & 0.0664 & 0.0662 & 0.1011 & 0.0747 \\
        Sub-05 & 0.0493 & 0.0515 & 0.0704 & 0.0975 & 0.0672 \\
        Sub-06 & 0.0690 & 0.0719 & 0.0498 & 0.0966 & 0.0718 \\
        Sub-07 & 0.1267 & 0.1300 & 0.0914 & 0.1312 & 0.1198 \\
        Sub-08 & 0.0718 & 0.0727 & 0.1038 & 0.1165 & 0.0912 \\
        Sub-09 & 0.0529 & 0.0563 & 0.0781 & 0.0756 & 0.0657 \\
        Sub-10 & 0.1122 & 0.1151 & 0.0961 & 0.1149 & 0.1096 \\
        \midrule
        \rowcolor{mylightgray} \textbf{Average} & 0.0810 & 0.0832 & 0.0838 & 0.1023 & 0.0876  \\
        \bottomrule
    \end{tabular}
\end{table}

Tab.~\ref{tab:mse_compact} presents the mean squared error (MSE) between the synthetic EEG signals generated by AlexNet and CORnet-S, and the real EEG signals for 10 subjects in THINGS-EEG2. The MSE was computed for each individual test sample and then averaged across the entire test set. Lower MSE values indicate better alignment between the synthetic and real EEG signals.

From the comparison shown in the Fig.~\ref{fig:appendix_retrieval_bar}, the retrieval accuracy for S-S (both training and testing sets consist of generated signals) is significantly higher than other categories, including T-T (both training and testing sets consist of real signals), T-S (training set consists of real signals, testing set consists of generated signals), and S-T (training set consists of generated signals, testing set consists of real signals), under both AlexNet and CORnet-S models. This indicates:

\textbf{Advantages of generated signals} Supported by black-box ANN models (e.g., AlexNet and CORnet-S), generated signals perform significantly better in retrieval tasks compared to real signals. In particular, the highest retrieval accuracy for S-S demonstrates the consistency and model adaptability of generated signals in this retrieval task. Different ANN models (e.g., AlexNet and CORnet-S) show consistent superiority in the retrieval tasks for generated signals, indicating that generated signals are more easily captured and distinguished by black-box proxy models.

\begin{figure}[htbp]
    \vspace{-20pt}
    \centering
    \includegraphics[width=0.9\textwidth]{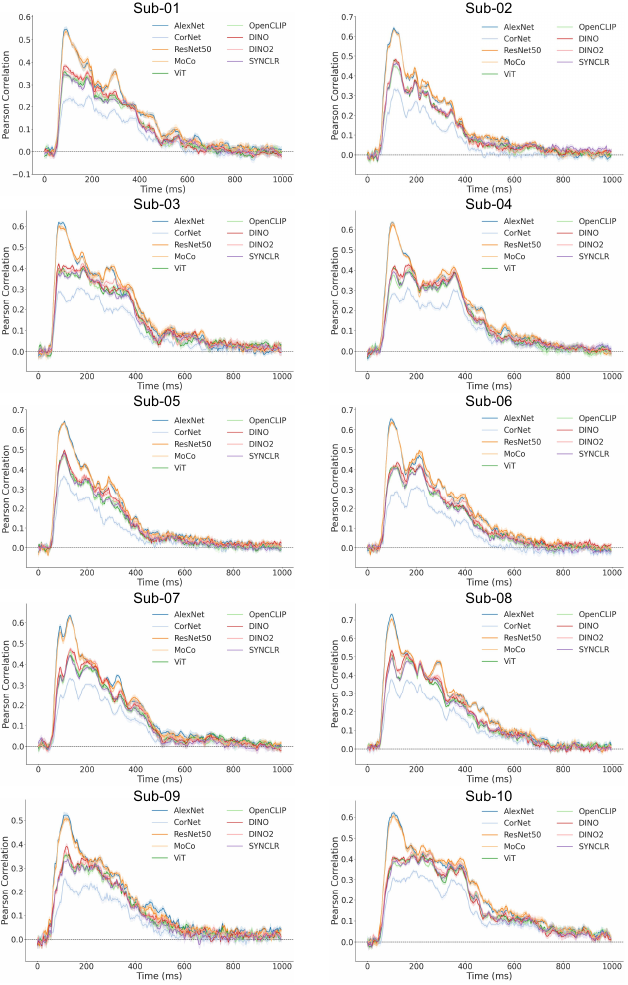}
    \caption{Time-resolved pearson correlation between ground truth and synthetic EEG signals. These synthetic signals were generated using a linearizing model fitted on feature maps from a range of pre-trained deep neural networks (AlexNet, CORnet-S, ResNet50, MoCo, ViT-B-32, OpenCLIP-ViT-B-32, DINO2-ViT-B-14, DINO-ViT-B-16, and SYNCLR-ViT-B-16).}
    \label{fig:pearson_timepoint}
\end{figure}

\paragraph{Correlation analysis} In Fig.~\ref{fig:pearson_timepoint}, for each time point, we compute the Pearson correlation between the real EEG signal and the synthetic signals. This analysis enables us to visualize how well each model replicates the temporal structure of real neural responses to visual stimuli, with shaded regions representing the standard deviation across samples. Notably, the results reveal that supervised models (e.g., ResNet) frequently outperform representation learning models (e.g., CLIP) in capturing these temporal dynamics. This contrasts with fMRI studies prioritizing semantic alignment~\citep{wang2023better}, likely due to the high temporal resolution of non-invasive EEG, which is inherently more sensitive to transient, low-to-mid-level visual features (e.g., edges, textures) during early processing stages~\citep{groen2017contributions}. ResNet, optimized for hierarchical feature extraction, aligns better with these "shallower" visual dynamics than the abstract semantic representations of CLIP. Furthermore, the dimensionality reduction (PCA) inherent to the encoding pipeline may favor the robust hierarchical features of ResNet over the complex joint-space representations of contrastive models for the specific task of Image-to-EEG synthesis.

\begin{figure}[ht]
    \vspace{-20pt}
    \centering
    \includegraphics[trim=0cm 2cm 0cm 0.5cm,clip,  width=0.95\textwidth]{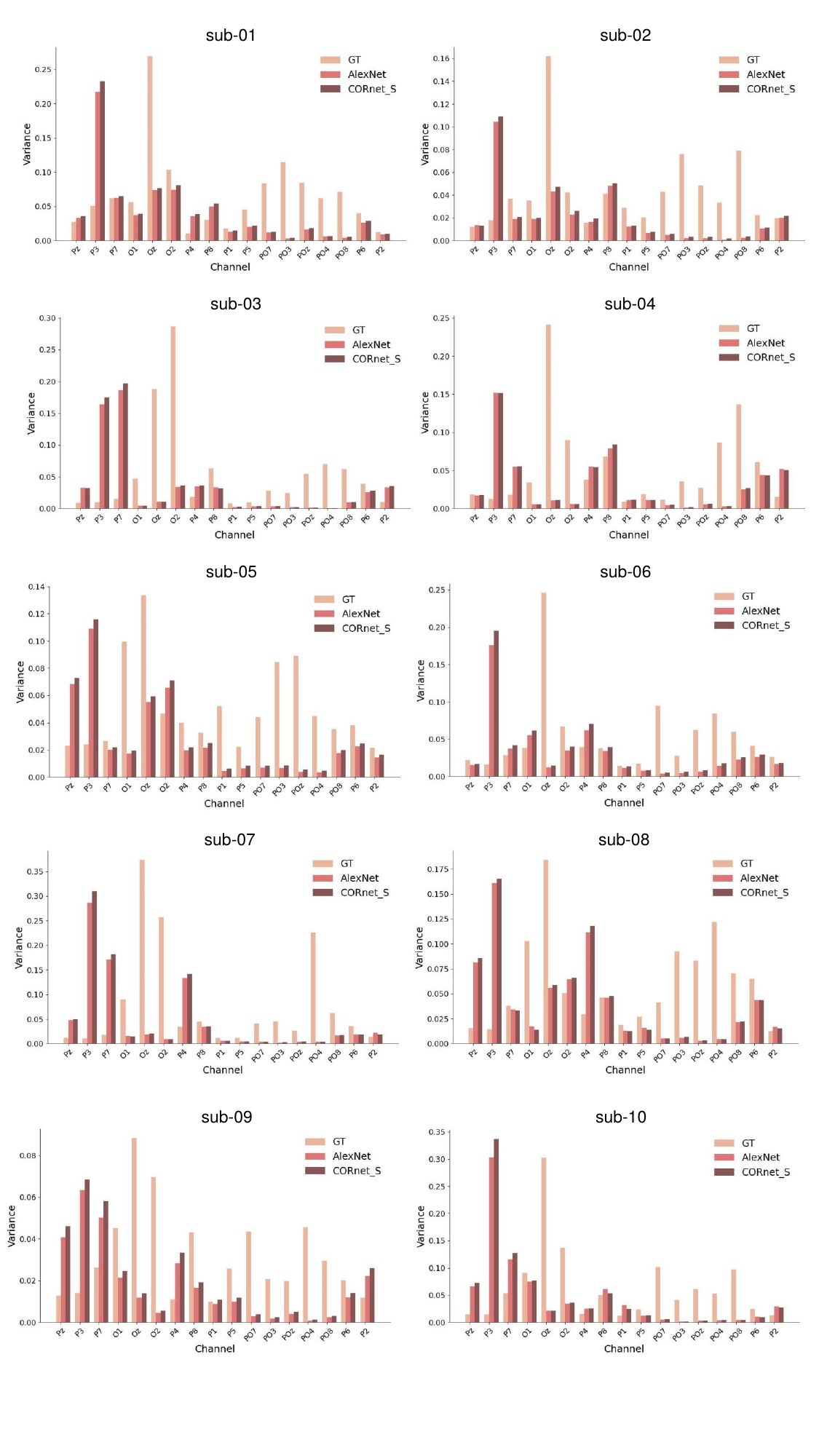}
    \caption{Variance across different channels for different visual stimulus and temporal dynamics}
    \label{fig:channel_var}
\end{figure}

\begin{figure}[ht]
    \vspace{-50pt}
    \centering
    \includegraphics[trim=0cm 2cm 0cm 2cm,clip,  width=0.95\textwidth]{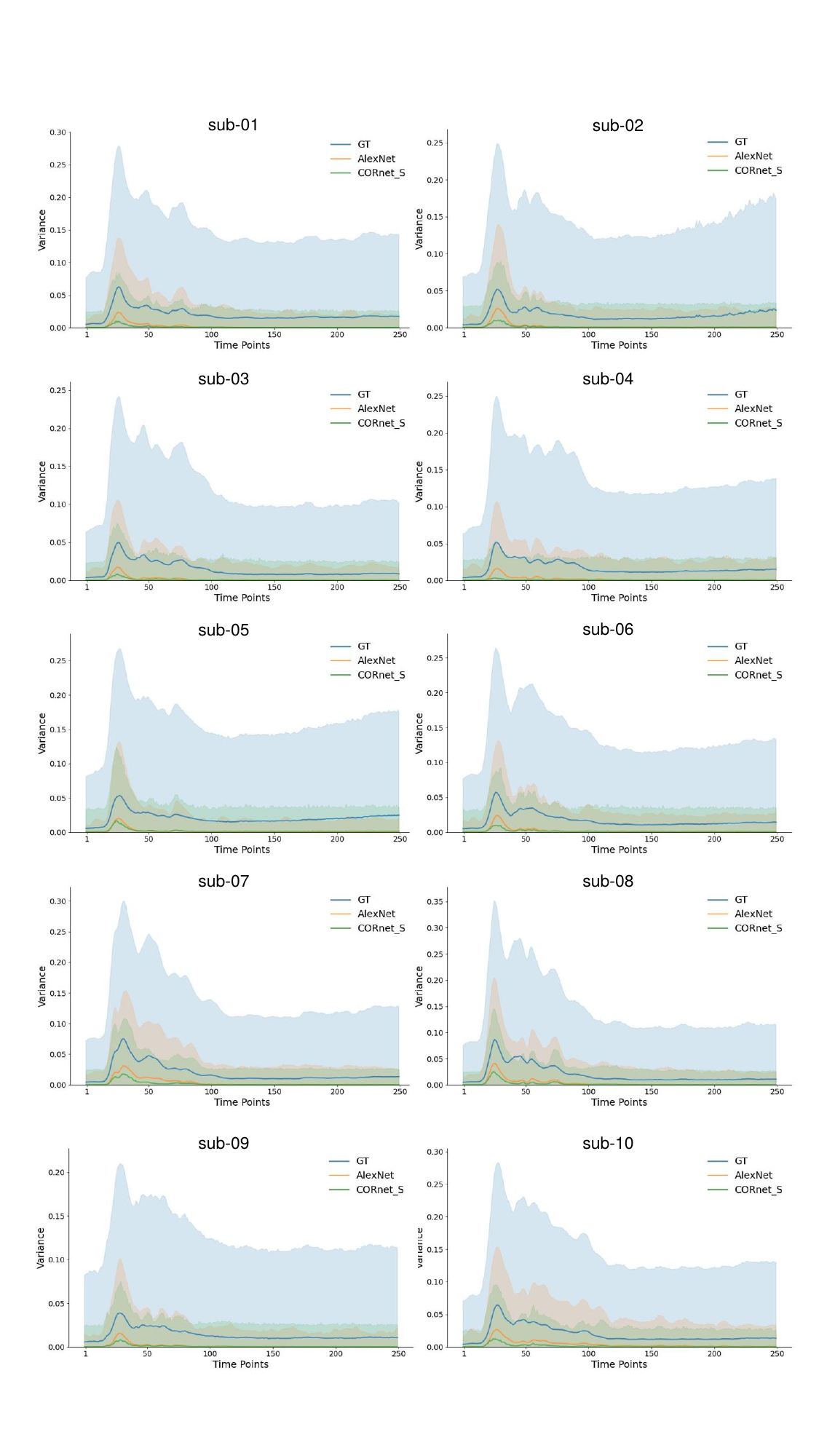}
    \caption{Variance across different time points for different visual stimuli and channels.}
    \label{fig:timepoint_var}
\end{figure}

\paragraph{Variance analysis} In Fig.~\ref{fig:channel_var}, we compute the variance across all samples and time points for each channel, providing a measure of the overall variability of the EEG signals in response to different visual stimuli and their temporal dynamics. This variance can help identify channels with the highest variability, which may be useful for selecting specific channels for further analysis or modulation. In Fig.~\ref{fig:timepoint_var}, we show the variance and standard deviation of the EEG signals computed across samples for each time point, and then averaged across channels. This analysis allows us to assess how signal variability evolves over time. By comparing the real EEG data with synthetic data generated by AlexNet and CORnet-S, we can evaluate how well each model captures the temporal variability present in the real EEG signals.

\begin{figure}[htbp]
    \vspace{-50pt}
    \centering
    \includegraphics[trim=0cm 3cm 0cm 0cm,clip,  width=0.95\textwidth]{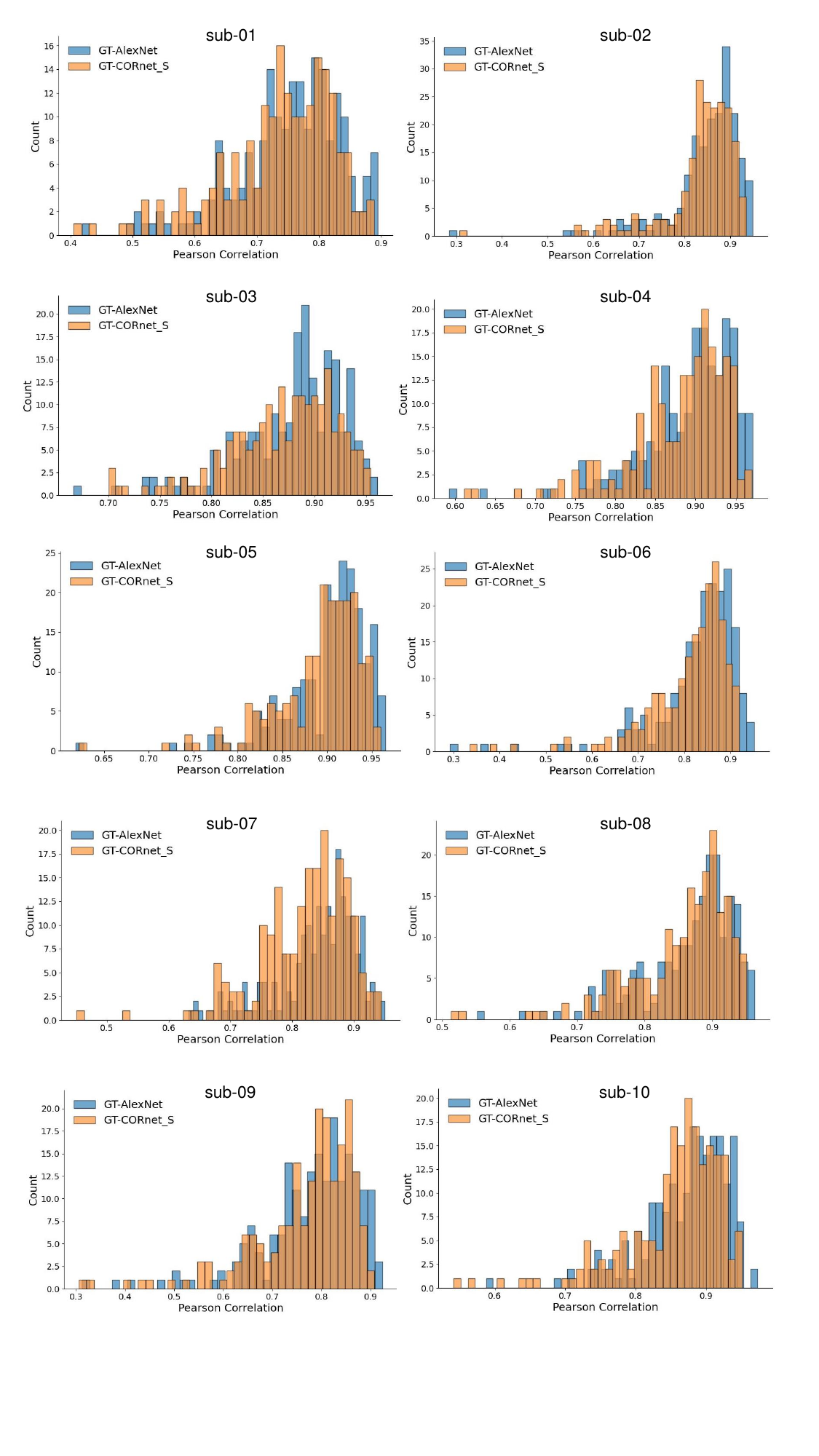}
    \caption{Distribution of pearson correlation coefficients across all sample pairs.}
    \label{fig:pearson_sample}
\end{figure}

\paragraph{Correlation distribution analysis} In Fig.~\ref{fig:pearson_sample}, we compute the pearson correlation coefficient between the averaged real EEG data and the synthetic data for each stimulus, measuring how well the synthetic data matches the real EEG on a per-sample basis. The histogram shows the distribution of correlation coefficients across all samples for both AlexNet and CORnet-S. A higher concentration of peaks near higher Pearson coefficients indicates better alignment between the synthetic data and the real EEG, reflecting superior model performance. 

These findings highlight the robustness of our EEG encoding models, demonstrating their ability that not only mimic the structural features of real EEG data but also capture the realistic variability seen in neural responses to visual stimuli. This suggests that our models are effective in approximating the neural representations underlying visual processing.

\clearpage
\subsection{Additional Interactive Searching Examples}
\label{sec:additional_retrieval_examples}
\subsubsection{More Examples of Interactive Searching}
\begin{figure}[htbp]
    \centering
    \includegraphics[width=0.9\linewidth]{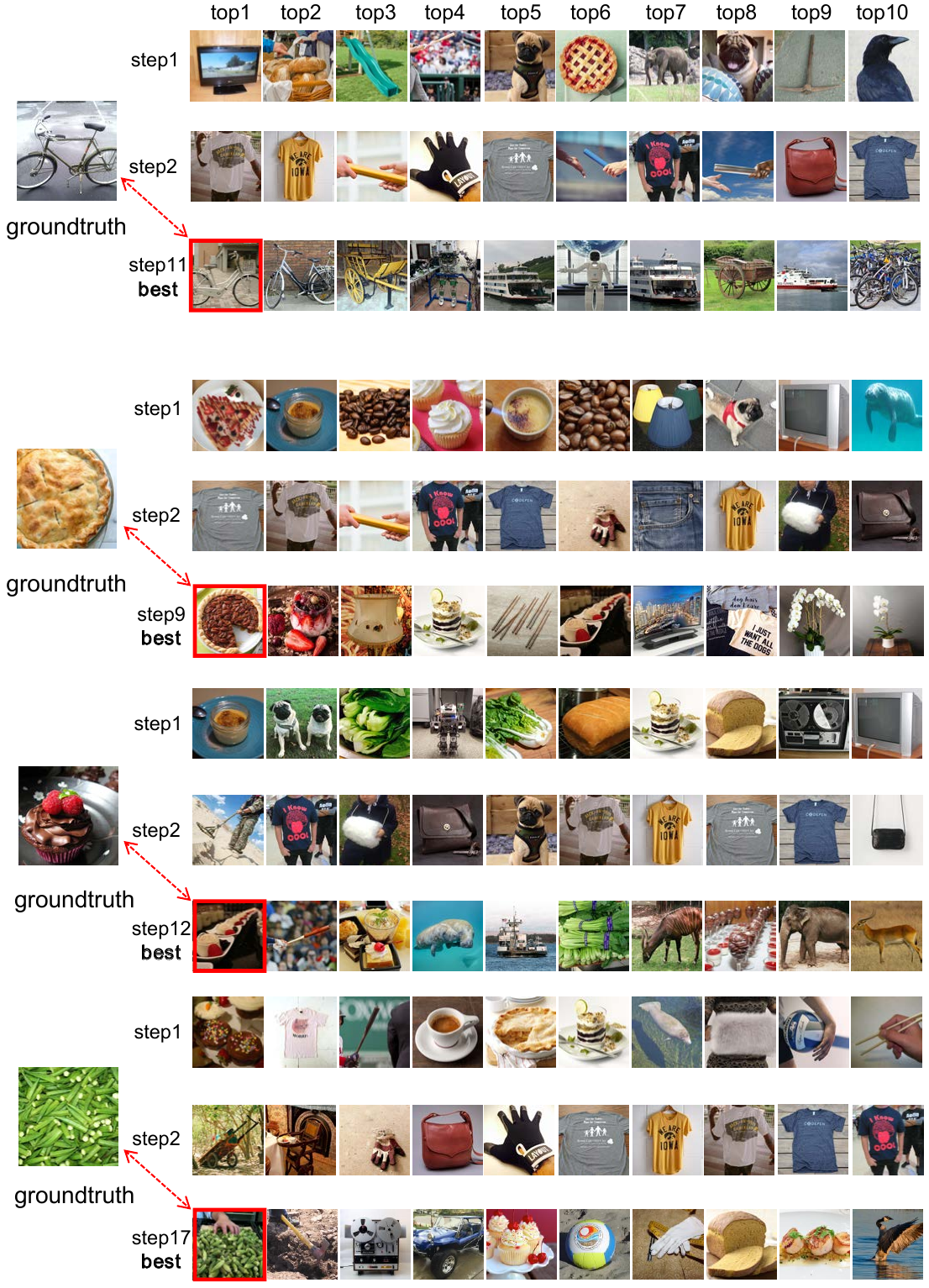}
    \caption{\textbf{Some searching examples of Subject 8, 4, 4, and 1.} By setting different targets, we present examples where the stimulus retrieved at the end of the iterative optimization process increasingly approximates the true category.}    
    \label{fig:appendix_correct_retrieval_examples}
\end{figure}

\subsubsection{Some Failure Examples of Interactive Searching}
\begin{figure}[htbp]
    \centering
    \includegraphics[width=1.0\linewidth]{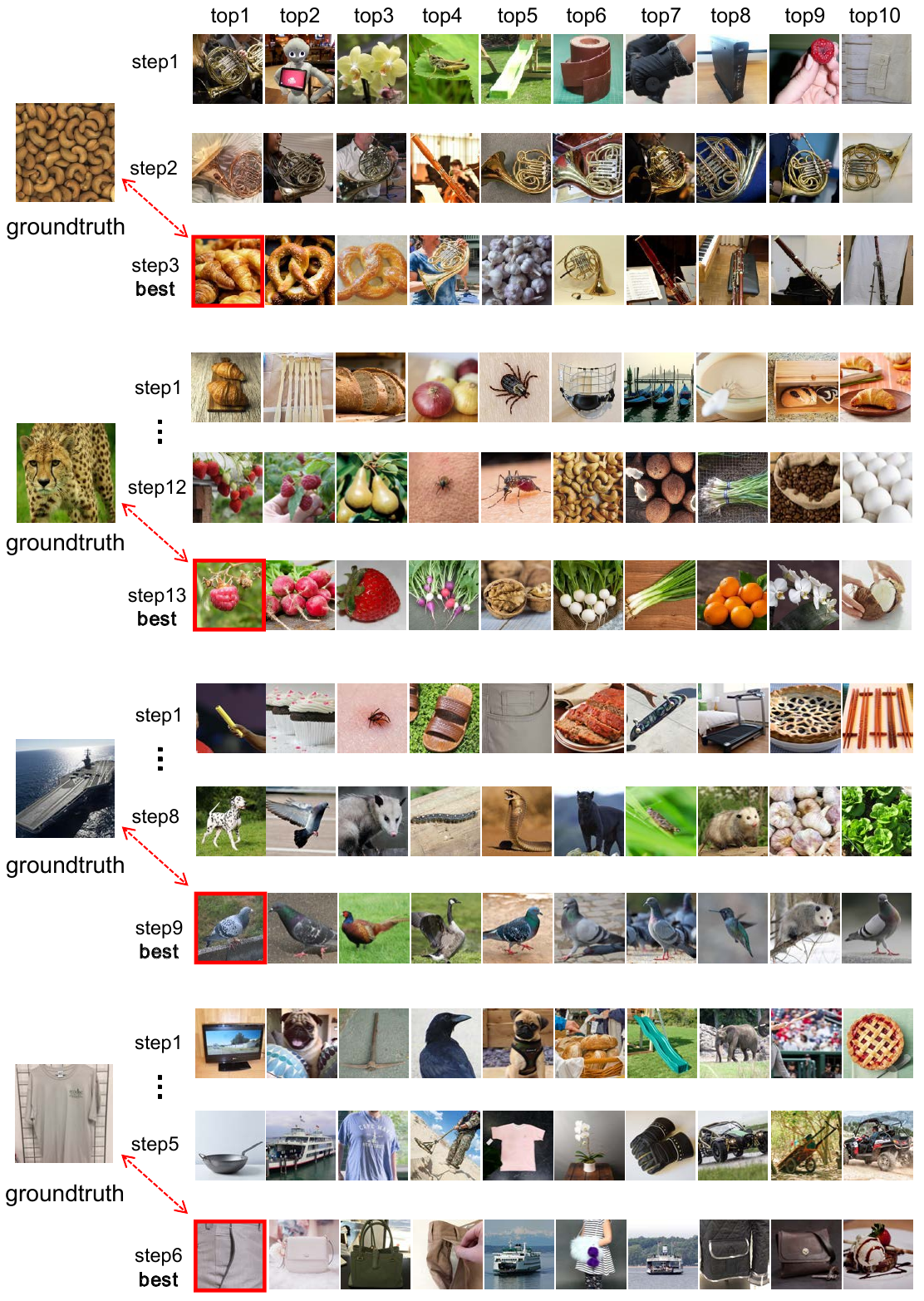}
    \caption{\textbf{Some failure examples from Subject 8.} By setting different targets, we show examples where the stimulus retrieved at the end of the iteration is far from the true category. In these examples, the final retrieved stimulus exhibits varying degrees of similarity to the target image.}    
    \label{fig:appendix_failure_retrieval_examples}
\end{figure}

\clearpage
\subsection{Additional Heuristic Generation Examples}
\label{sec:additional_generation_examples}

\begin{figure}[htbp]
    \centering
    \includegraphics[width=0.75\textwidth]{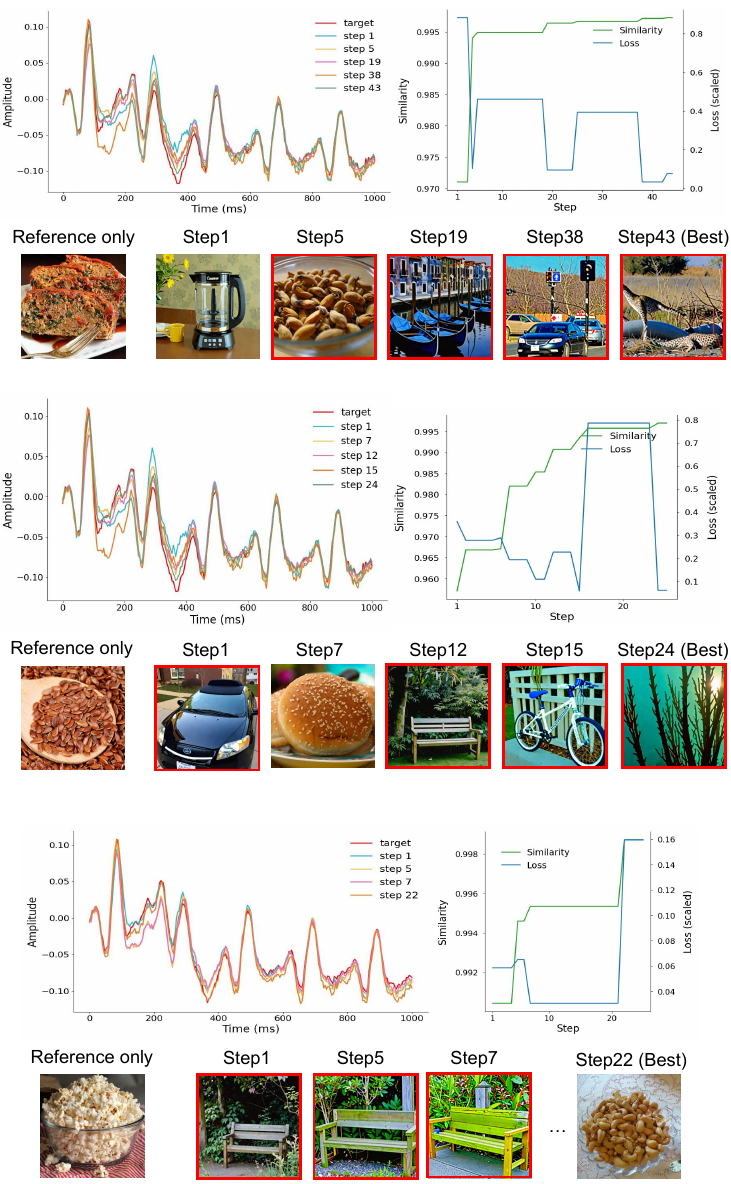}
    \caption{\textbf{Illustration of the closed-loop iterative process for Subject 1.} Three distinct visual targets were presented, each based on a specific similarity measure (details in Target Features of EEG, Section~\ref{target}), with new visual stimuli iteratively generated for each target. The left panel illustrates the time-domain evolution of neural responses across iterations. The right panel depicts the changes in similarity (green curve) and loss (blue curve, scaled) between the current stage features and the target features.}    
    \label{fig:appendix_generation_examples_sub01}
\end{figure}

\begin{figure}[htbp]
    \centering
    \includegraphics[width=0.8\textwidth]{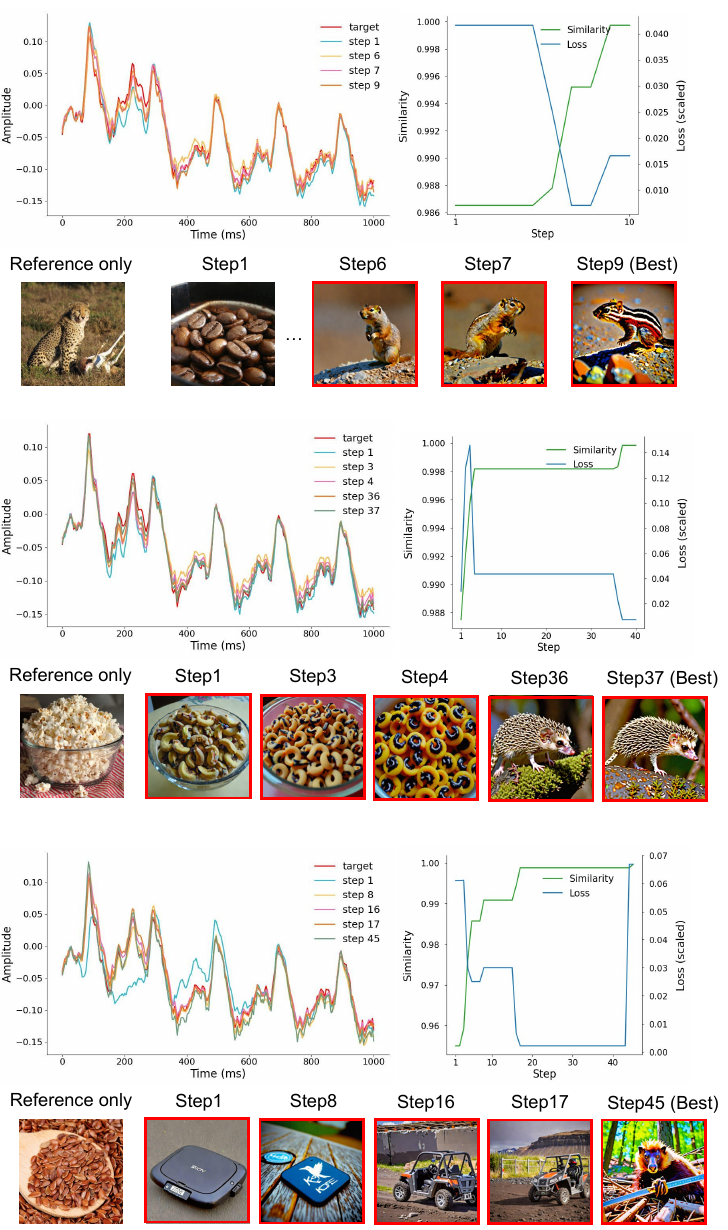}
    \caption{\textbf{Illustration of the closed-loop iterative process for Subject 2.} Three distinct visual targets were presented, each based on a specific similarity measure (details in Target Features of EEG, Section~\ref{target}), with new visual stimuli iteratively generated for each target. The left panel illustrates the time-domain evolution of neural responses across iterations. The right panel depicts the changes in similarity (green curve) and loss (blue curve, scaled) between the current stage features and the target features.}    
    \label{fig:appendix_generation_examples_sub02}
\end{figure}

\begin{figure}[htbp]
    \centering
    \includegraphics[width=0.8\textwidth]{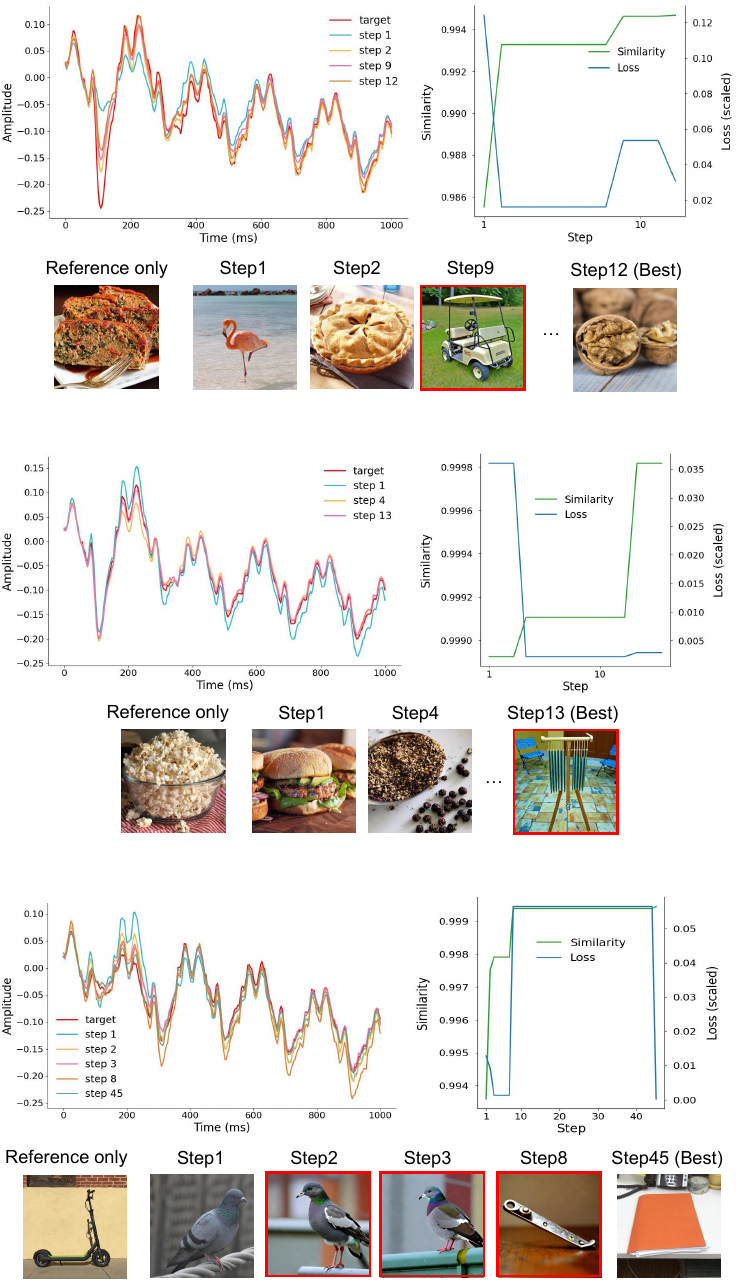}
    \caption{\textbf{Illustration of the closed-loop iterative process for Subject 3.} Three distinct visual targets were presented, each based on a specific similarity measure (details in Target Features of EEG, Section~\ref{target}), with new visual stimuli iteratively generated for each target. The left panel illustrates the time-domain evolution of neural responses across iterations. The right panel depicts the changes in similarity (green curve) and loss (blue curve, scaled) between the current stage features and the target features.}    
    \label{fig:appendix_generation_examples_sub03}
\end{figure}

\begin{figure}[htbp]
    \centering
    \includegraphics[width=0.8\textwidth]{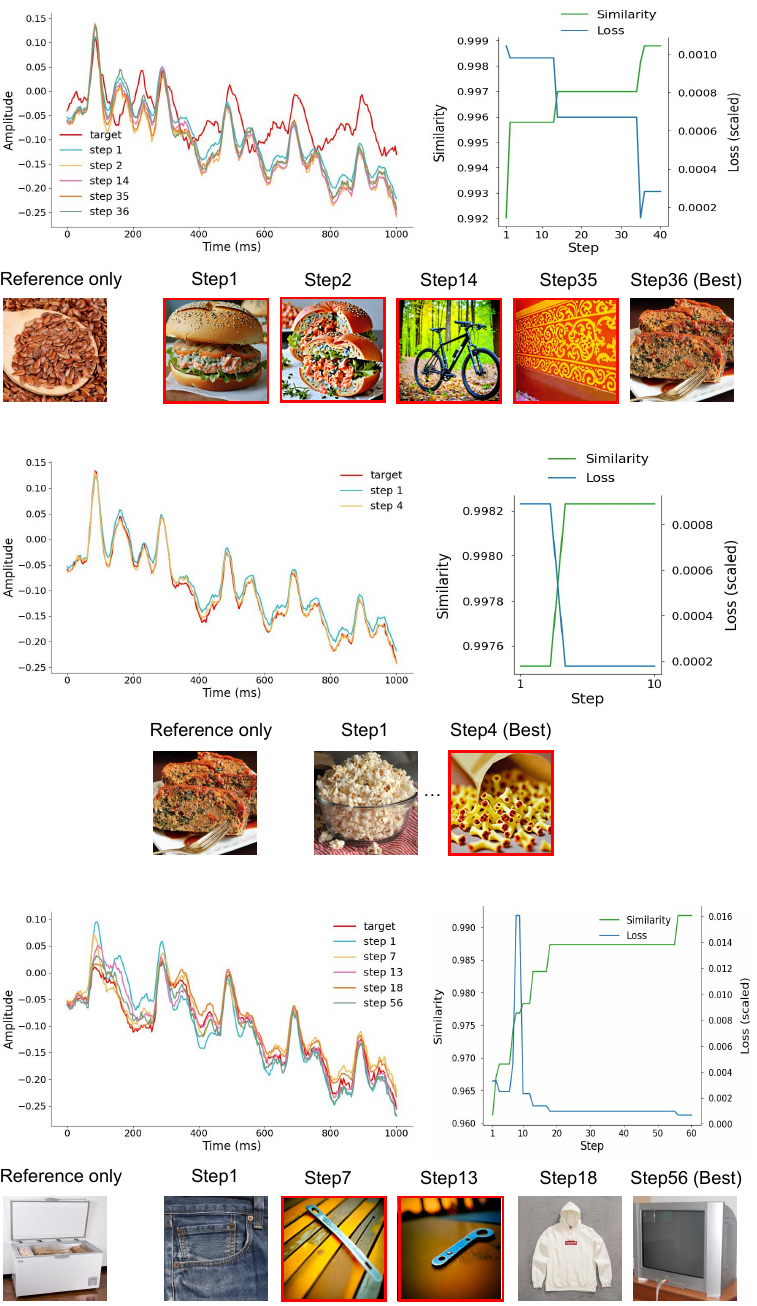}
    \caption{\textbf{Illustration of the closed-loop iterative process for Subject 4.} Three distinct visual targets were presented, each based on a specific similarity measure (details in Target Features of EEG, Section~\ref{target}), with new visual stimuli iteratively generated for each target. The left panel illustrates the time-domain evolution of neural responses across iterations. The right panel depicts the changes in similarity (green curve) and loss (blue curve, scaled) between the current stage features and the target features.}    
    \label{fig:appendix_generation_examples_sub04}
\end{figure}

\begin{figure}[htbp]
    \centering
    \includegraphics[width=0.8\textwidth]{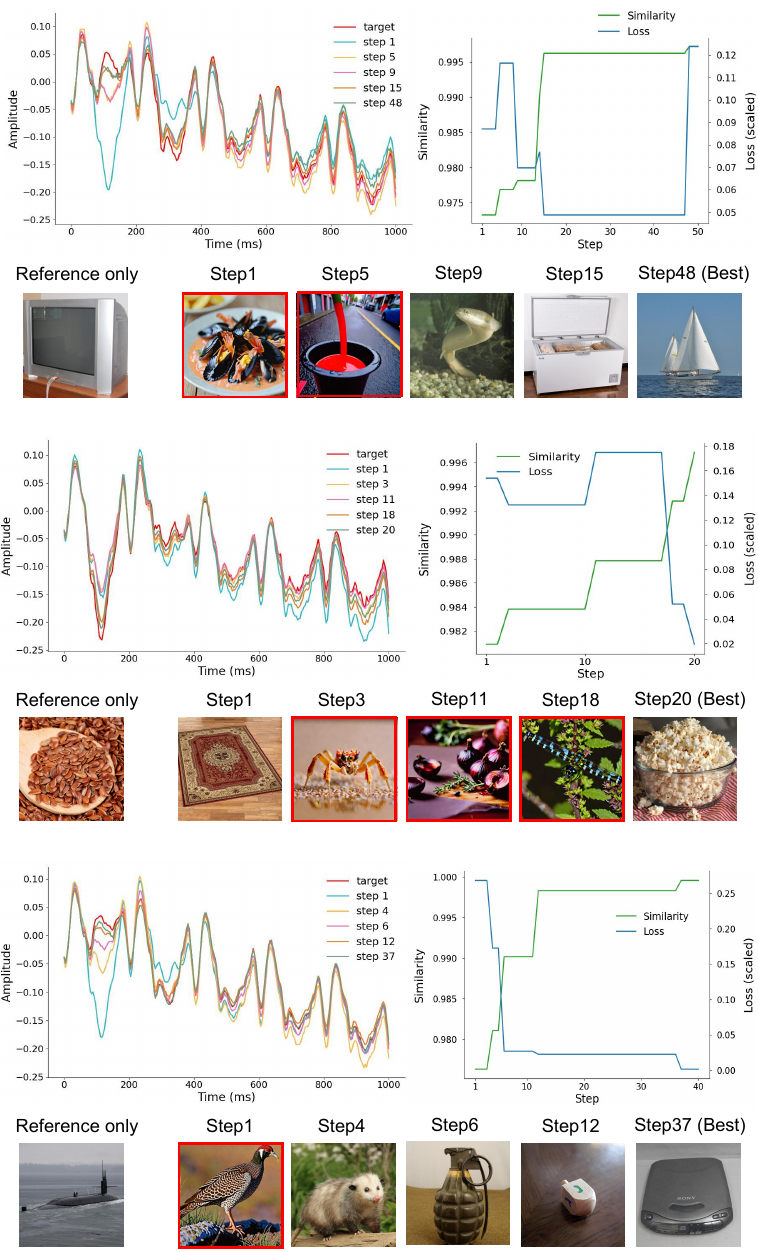}
    \caption{\textbf{Illustration of the closed-loop iterative process for Subject 5.} Three distinct visual targets were presented, each based on a specific similarity measure (details in Target Features of EEG, Section~\ref{target}), with new visual stimuli iteratively generated for each target. The left panel illustrates the time-domain evolution of neural responses across iterations. The right panel depicts the changes in similarity (green curve) and loss (blue curve, scaled) between the current stage features and the target features.}    
    \label{fig:appendix_generation_examples_sub05}
\end{figure}

\begin{figure}[htbp]
    \centering
    \includegraphics[width=0.8\textwidth]{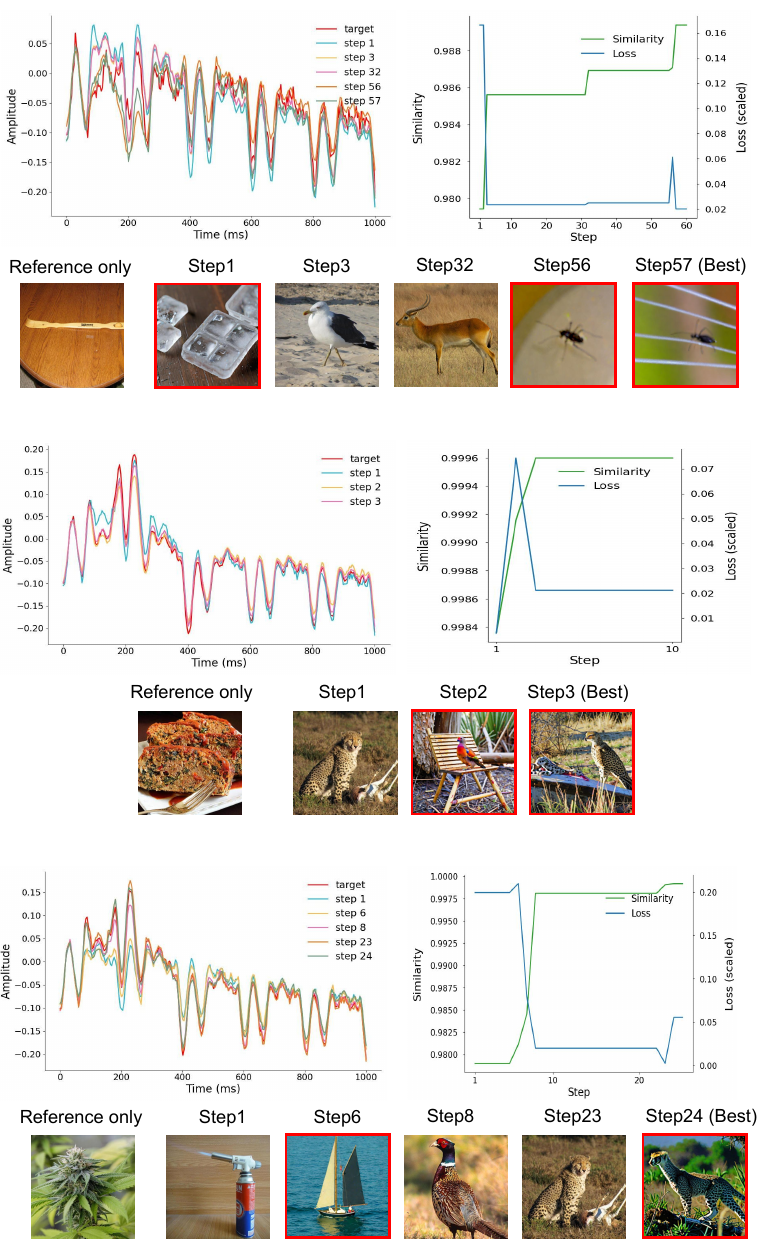}
    \caption{\textbf{Illustration of the closed-loop iterative process for Subject 6.} Three distinct visual targets were presented, each based on a specific similarity measure (details in Target Features of EEG, Section~\ref{target}), with new visual stimuli iteratively generated for each target. The left panel illustrates the time-domain evolution of neural responses across iterations. The right panel depicts the changes in similarity (green curve) and loss (blue curve, scaled) between the current stage features and the target features.}    
    \label{fig:appendix_generation_examples_sub06}
\end{figure}

\begin{figure}[htbp]
    \centering
    \includegraphics[width=0.8\textwidth]{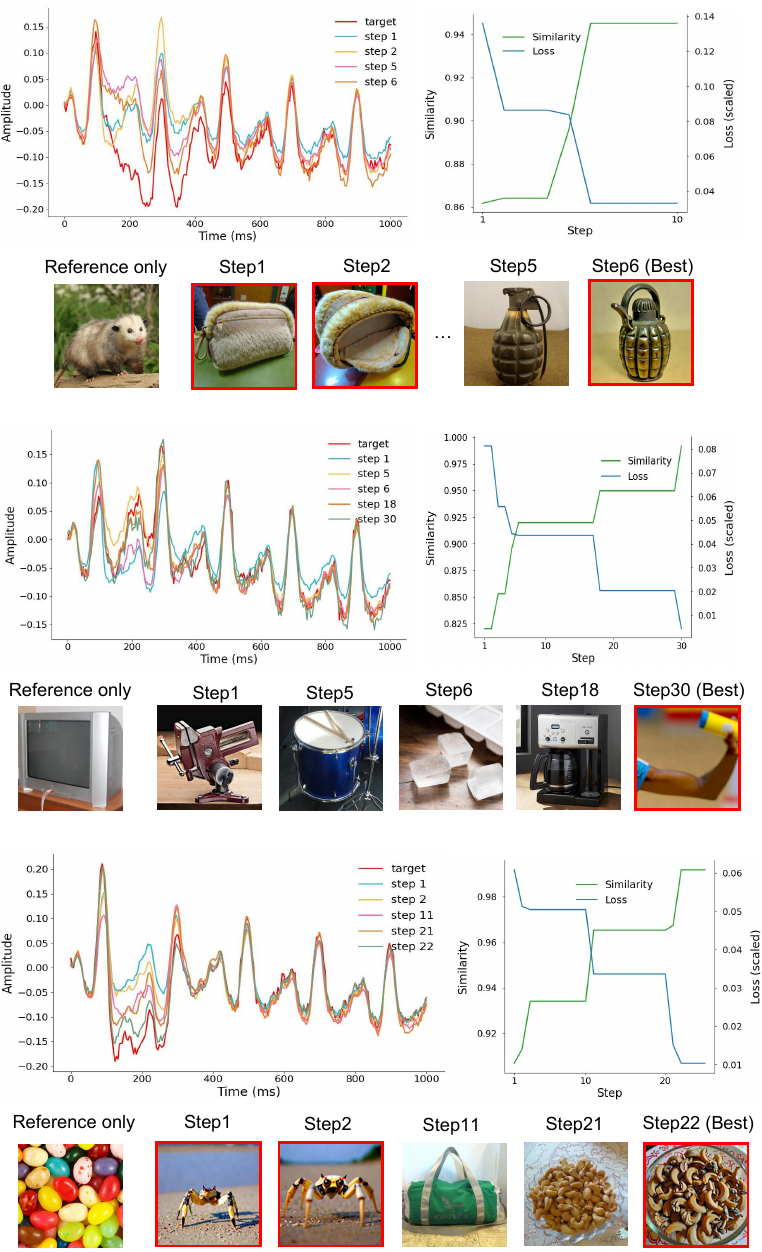}
    \caption{\textbf{Illustration of the closed-loop iterative process for Subject 7.} Three distinct visual targets were presented, each based on a specific similarity measure (details in Target Features of EEG, Section~\ref{target}), with new visual stimuli iteratively generated for each target. The left panel illustrates the time-domain evolution of neural responses across iterations. The right panel depicts the changes in similarity (green curve) and loss (blue curve, scaled) between the current stage features and the target features.}    
    \label{fig:appendix_generation_examples_sub07}
\end{figure}

\begin{figure}[htbp]
    \centering
    \includegraphics[width=0.8\textwidth]{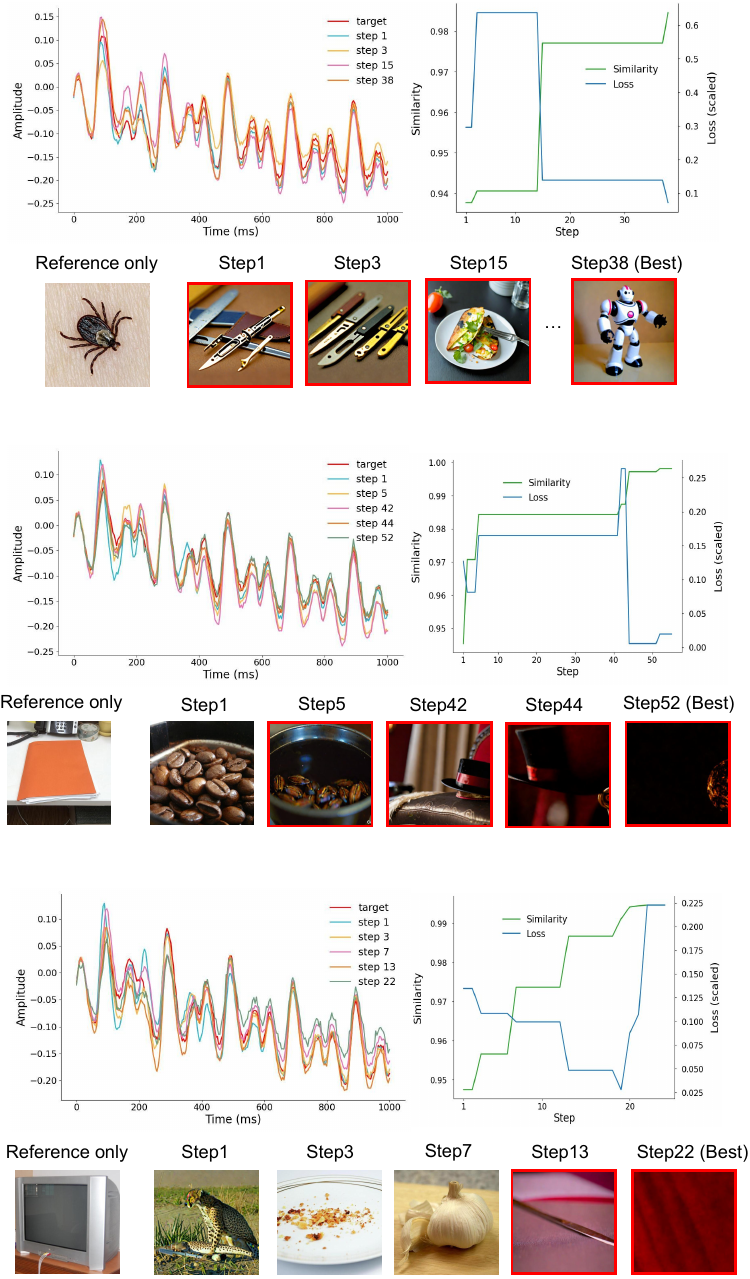}
    \caption{\textbf{Illustration of the closed-loop iterative process for Subject 8.} Three distinct visual targets were presented, each based on a specific similarity measure (details in Target Features of EEG, Section~\ref{target}), with new visual stimuli iteratively generated for each target. The left panel illustrates the time-domain evolution of neural responses across iterations. The right panel depicts the changes in similarity (green curve) and loss (blue curve, scaled) between the current stage features and the target features.}    
    \label{fig:appendix_generation_examples_sub08}
\end{figure}

\begin{figure}[htbp]
    \centering
    \includegraphics[width=0.8\textwidth]{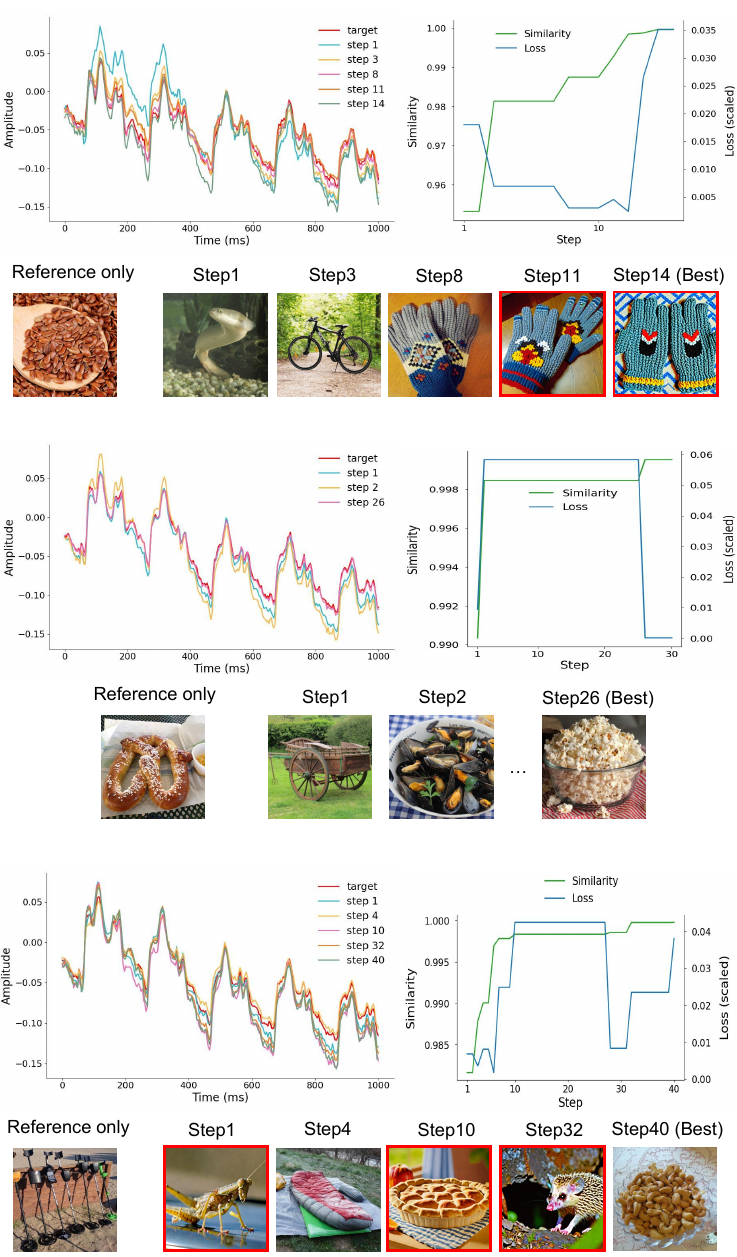}
    \caption{\textbf{Illustration of the closed-loop iterative process for Subject 9.} Three distinct visual targets were presented, each based on a specific similarity measure (details in Target Features of EEG, Section~\ref{target}), with new visual stimuli iteratively generated for each target. The left panel illustrates the time-domain evolution of neural responses across iterations. The right panel depicts the changes in similarity (green curve) and loss (blue curve, scaled) between the current stage features and the target features.}    
    \label{fig:appendix_generation_examples_sub9}
\end{figure}

\begin{figure}[htbp]
    \centering
    \includegraphics[width=0.8\textwidth]{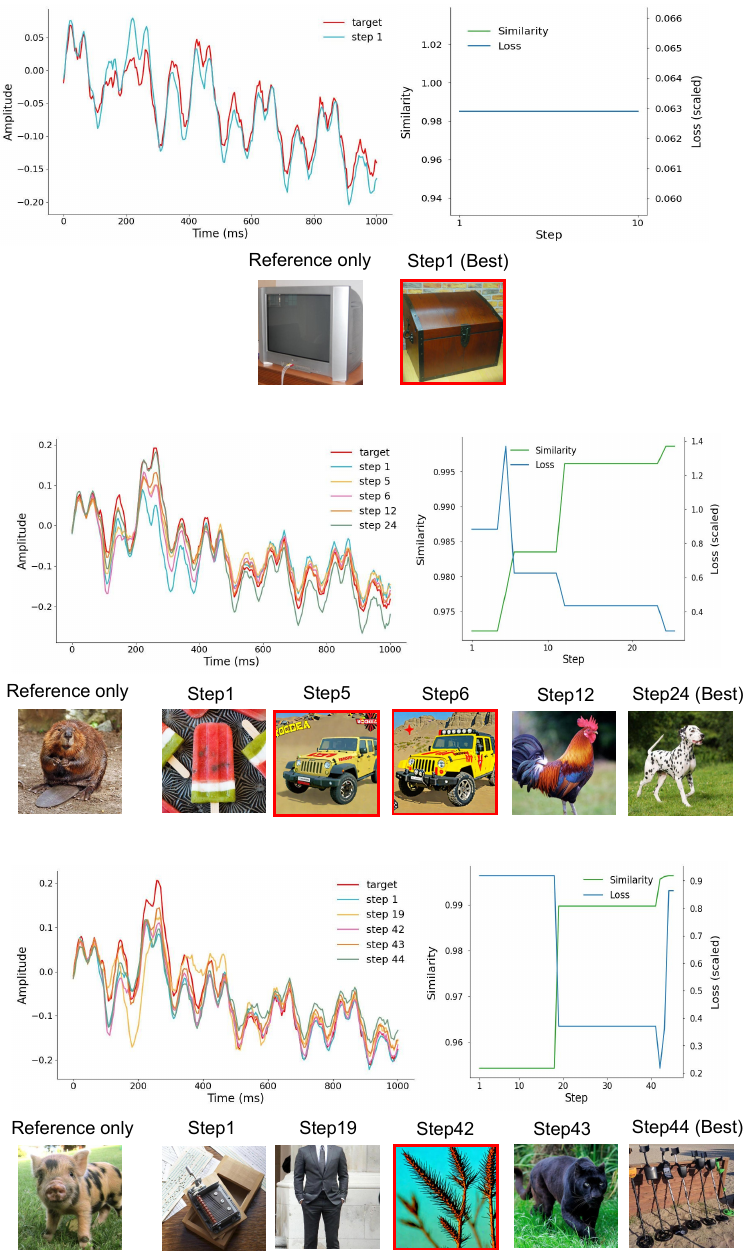}
    \caption{\textbf{Illustration of the closed-loop iterative process for Subject 10.} Three distinct visual targets were presented, each based on a specific similarity measure (details in Target Features of EEG, Section~\ref{target}), with new visual stimuli iteratively generated for each target. The left panel illustrates the time-domain evolution of neural responses across iterations. The right panel depicts the changes in similarity (green curve) and loss (blue curve, scaled) between the current stage features and the target features.}    
    \label{fig:appendix_generation_examples_sub10}
\end{figure}

\end{document}